\documentclass[Afour,sagev,times]{sagej}

\usepackage{moreverb,url}
\usepackage{bm}
\usepackage[colorlinks,bookmarksopen,bookmarksnumbered,citecolor=red,urlcolor=red]{hyperref}
\usepackage{subcaption}
\usepackage{floatrow}
\newfloatcommand{capbtabbox}{table}[][\FBwidth]
\usepackage{booktabs}
\usepackage{algorithm}
\usepackage{algpseudocode}

\newcommand\BibTeX{{\rmfamily B\kern-.05em \textsc{i\kern-.025em b}\kern-.08em
T\kern-.1667em\lower.7ex\hbox{E}\kern-.125emX}}

\def\volumeyear{2022}

\begin{document}

\runninghead{Liu et al.}

\title{Neural Extended Kalman Filters for Learning and Predicting Dynamics of Structural Systems}

\author{Wei Liu\affilnum{1,2}, Zhilu Lai\affilnum{2,3,4,5}, Kiran Bacsa\affilnum{2,3} and Eleni Chatzi\affilnum{2,3}}

\affiliation{\affilnum{1}Department of Industrial Systems Engineering and Management, National University of Singapore, Singapore, Singapore\\
\affilnum{2}Chair of Structural Mechanics and Monitoring, Department of Civil, Environmental and Geomatic Engineering, ETH Z\"urich, Z\"urich, Switzerland\\
\affilnum{3}Future Resilient Systems, Singapore-ETH Centre, Singapore 138602, Singapore\\
\affilnum{4}Internet of Things Thrust, Information Hub, HKUST(GZ), Guangzhou, China\\
\affilnum{5}Department of Civil and Environmental Engineering, HKUST, Hong Kong, China
}

\corrauth{Zhilu Lai, Internet of Things Thrust, Information Hub, HKUST(GZ), Guangzhou 511458, China. Formerly at Singapore-ETH Centre.}

\email{zhilulai@ust.hk}

\begin{abstract}
Accurate structural response prediction forms a main driver for structural health monitoring and control applications. This often requires the proposed model to adequately capture the underlying dynamics of complex structural systems. In this work, we utilize a learnable Extended Kalman Filter (EKF), named the Neural Extended Kalman Filter (Neural EKF) throughout this paper, for learning the latent evolution dynamics of complex physical systems. The Neural EKF is a generalized version of the conventional EKF, where the modeling of process dynamics and sensory observations can be parameterized by neural networks, therefore learned by end-to-end training. The method is implemented under the variational inference framework with the EKF conducting inference from sensing measurements. Typically, conventional variational inference models are parameterized by neural networks independent of the latent dynamics models. This characteristic makes the inference and reconstruction accuracy weakly based on the dynamics models and renders the associated training inadequate. In this work, we show that the structure imposed by the Neural EKF is beneficial to the learning process. We demonstrate the efficacy of the framework on both simulated and real-world structural monitoring datasets, with the results indicating significant predictive capabilities of the proposed scheme.
\end{abstract}


\maketitle

\section{Introduction}
Digital twins have attracted interest in a wide range of applications, among which structural digital twins have shown promise in virtual health monitoring, decision-making and predictive maintenance \cite{tao2018digital}. A digital twin is a virtual representation of a connected physical asset and encompasses its entire product lifecycle \cite{aiaa2020digital}. They acquire and assimilate observational data from the physical system and adopt this information to update dynamics models, which represent the evolving physical system \cite{kapteyn2021probabilistic}. The underlying dynamic model thus significantly underpins the establishment of a digital twin \cite{wagg2020digital}. Unlike physics-based modeling, which relies on first principles \cite{kerschen2006past}, data-driven modeling resorts to use of learning algorithms to infer the underlying mechanisms that drive a system’s response \cite{brunton2022data}. The latter class of models plays an essential role in vibration-based structural health monitoring (SHM) \cite{farrar2012structural}, as a main approach to understanding the actual properties of the assessed structural systems, and to performing accurate dynamical response prediction with the established models.
Diverse data-driven techniques
have been developed in the context of SHM, including vibration-based modeling based on modal parameters derived from measured structural response signals \cite{dohler2013uncertainty} and conventional time series analysis techniques, including those of the AutoRegressive class \cite{bogoevska2017data,avendano2021virtual}. These methods are typically non-trivial to apply on complex structures with large inherent uncertainties, such as large-scale civil infrastructure systems.

Further to classical data-driven approaches, which draw from system identification, machine learning (ML) and deep learning (DL), in particular, offer powerful and promising tools for modeling complex systems or capturing latent phenomena. Important applications include image recognition \cite{he2016deep,svensen2018deep,krizhevsky2012imagenet}, speech recognition \cite{hinton2012deep}, autonomous driving \cite{bojarski2016end}, and more. Their aptitude in recognizing patterns in data and doing so via use of reduced latent features renders such schemes potent tools for large complex structural dynamics modeling \cite{ljung2020deep,farrar2012structural,dervilis2013machine}, which exhibit complex, oftentimes nonlinear, behaviour.

Many recent works have been focusing on bridging deep learning and structural dynamics \cite{chen2018neural,jiang2017fuzzy}. We classify these methods into two main categories, namely direct mappings and generative models. The first class essentially learns a function that maps the input (force/load) to the output (response) of the system. This class of methods exploits neural networks with customized architectures and physical constraints tailored to the problem at hand \cite{raissi2019physics,zhang2019deep,zhang2020physics,xu2021phymdan}, such as recurrent neural networks (RNN) and convolutional neural networks (CNN).
The caveat of these methods lies in constructing a mere input-output mapping without explicitly modeling the governing dynamics, thus implying that interpretability and generalizability may not be guaranteed. Meanwhile, vibration data measured from operating engineered systems usually embody and reflect the underlying governing dynamics of that system.
In this respect, the second class of methods, which fall in the generative model class, attempts to learn a dynamical model that best captures the underlying dynamic evolution. Deep learning-based state space modeling, either in continuous or in discretized format, has been employed to perform data-driven modeling with various deep learning methods \cite{liu2022physics,lai2021structural,fraccaro2017disentangled,karl2016deep,gedon2020deep,girin2020dynamical, bacsa2023symplectic}. This class of methods aligns with the common approach to modeling dynamics, rendering such a representation interpretable, even though it is delivered via learned neural network functions. \textit{This paper will be focusing on the second class}. 
\begin{figure}[h]
\centering
\includegraphics[width=0.94\linewidth]{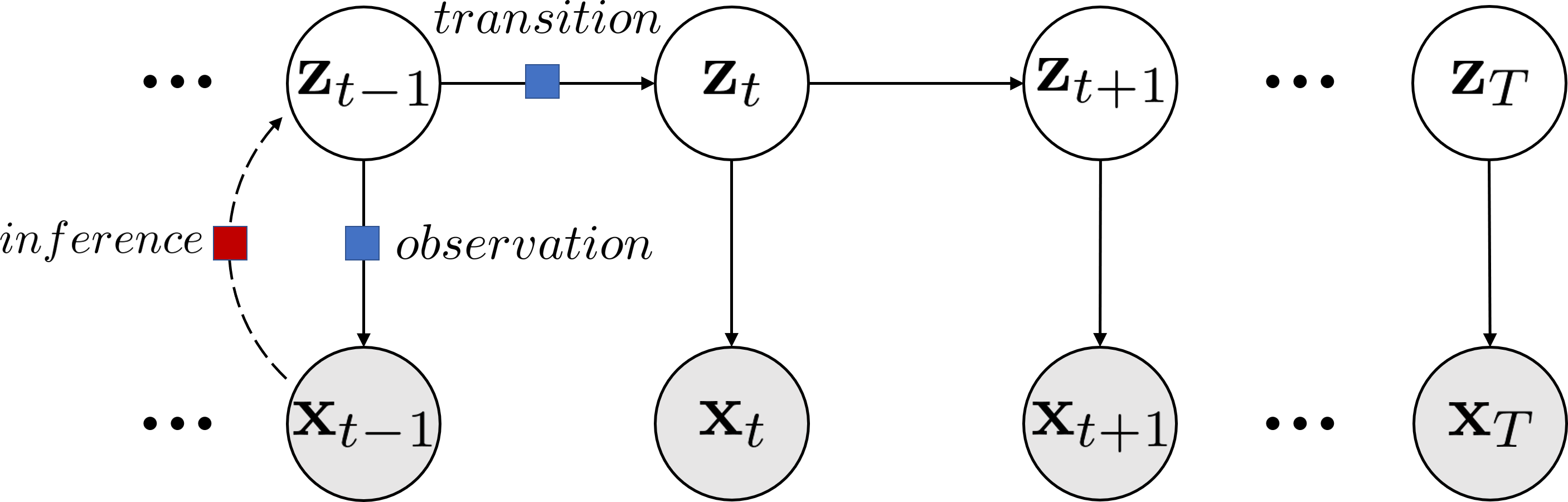}
\caption{Learning a dynamical system from data using a DL-based state space approach.}
\label{SSM}
\end{figure}

As shown in Figure \ref{SSM}, the key idea of this second class lies in learning three models: (i) the \textit{inference} model: modeling how the latent variable vector $\textbf{z}$ can be estimated from observation/data $\textbf{x}$; (ii) the \textit{transition} model: modeling how the latent variable $\textbf{z}$ evolves over time; and (iii) the \textit{observation} model: modeling how the measured data $\textbf{x}$ are generated from $\textbf{z}$. For convenience, the transition and observation models are collectively referred to as dynamics models throughout this paper. 


Dynamical variational autoencoders (VAEs) \cite{kingma2013auto,girin2020dynamical} is a class of methods following the architecture described in Figure \ref{SSM},  for modeling a dynamical system from data ($\textbf{x} \rightarrow \textbf{z} \rightarrow \textbf{x}$).
In addition to the dynamics models, the inference models in VAE-based methods are typically parameterized by neural networks. As shown in \cite{krishnan2017structured,karl2016deep} and studied further in \cite{li2021replay}, in this setting, the variational objective depends mainly on the inference model, while the dynamic models are relegated to a regularizer for the inference network. Often, this makes the learned model unsuitable for prediction due to the inaccuracy of the dynamics models employed to generate predictions. 

On the other hand, while VAEs can be too flexible with both dynamics models and inference models parameterized by neural networks, the conventional Bayesian filtering methods \cite{chatzi2009unscented,azam2015dual,chatzis2017discontinuous} go to the other extreme, requiring known dynamics models, usually derived from physics-based representations, and conduct the inference only using these dynamics models with no further inference models. Formulating suitable transition and observation models can be infeasible and challenging in many scenarios, especially when the dynamics of the system is not easily modeled or fully understood, or the sensory observations stem from vision-based data (images, videos). Over the past years, deep learning has become the method of choice for processing such data. Recent
work \cite{haarnoja2016backprop,li2021replay,kloss2021train,jonschkowski2018differentiable,karkus2018particle} showed that it is also possible to render the dynamics models of Bayesian filters learnable and train them end-to-end, leveraging deep learning techniques and filtering algorithms. For Kalman Filters \cite{haarnoja2016backprop}, Extended Kalman Filters \cite{li2021replay}, Unscented Kalman Filters \cite{kloss2021train} and Particle Filters \cite{jonschkowski2018differentiable,karkus2018particle},
the respective works showed that such learnable filters systematically outperform unstructured neural networks like RNN, LSTM and CNN \cite{kloss2021train}.

Combining the benefits of the VAE- and Kalman filtering-based strategies, and inspired by related work in the robotics community \cite{haarnoja2016backprop,li2021replay,kloss2021train,jonschkowski2018differentiable,karkus2018particle}, we utilize the learnable EKF proposed in \cite{li2021replay} and name it as \textit{Neural Extended Kalman Filter (Neural EKF)} in this paper for learning and predicting complex structural dynamics in real scenarios, based on these main principles: (i) we use EKF formulas to conduct closed-form inference instead of implementing an additional neural network to overtake this task,
thus allowing for the training to focus on the dynamics models rather than the inference task; and (ii) we eliminate the requirement of explicitly known transition and observation models imposed by conventional Bayesian filtering approaches. At this point, we wish to clarify that the idea we here present is originally adapted from \cite{li2021replay}, where a \textit{Replay Overshooting} framework is presented. The Neural EKF presented in this paper is a modified version of the previous scheme, adapted to structural dynamics problems. We further wish to clarify that an existing framework termed Neural EKF \cite{owen2003neural} appears in existing literature, which is however different to what we here propose. In that work, the transition model comprises the Jacobian of estimated system dynamics as a prior, with an error correction term modeled by neural networks. That framework is termed Neural EKF due to the use of the Jacobian in the transition model, while the neural network simply learns a discrepancy term. There is no actual extended Kalman filtering process, while further the observation and inference models are not considered, as carried out in this work. Here we choose to term this framework the Neural EKF since our suggested scheme comprises a fully learnable version of the original EKF.

\begin{figure}
\centering
\includegraphics[width=\linewidth]{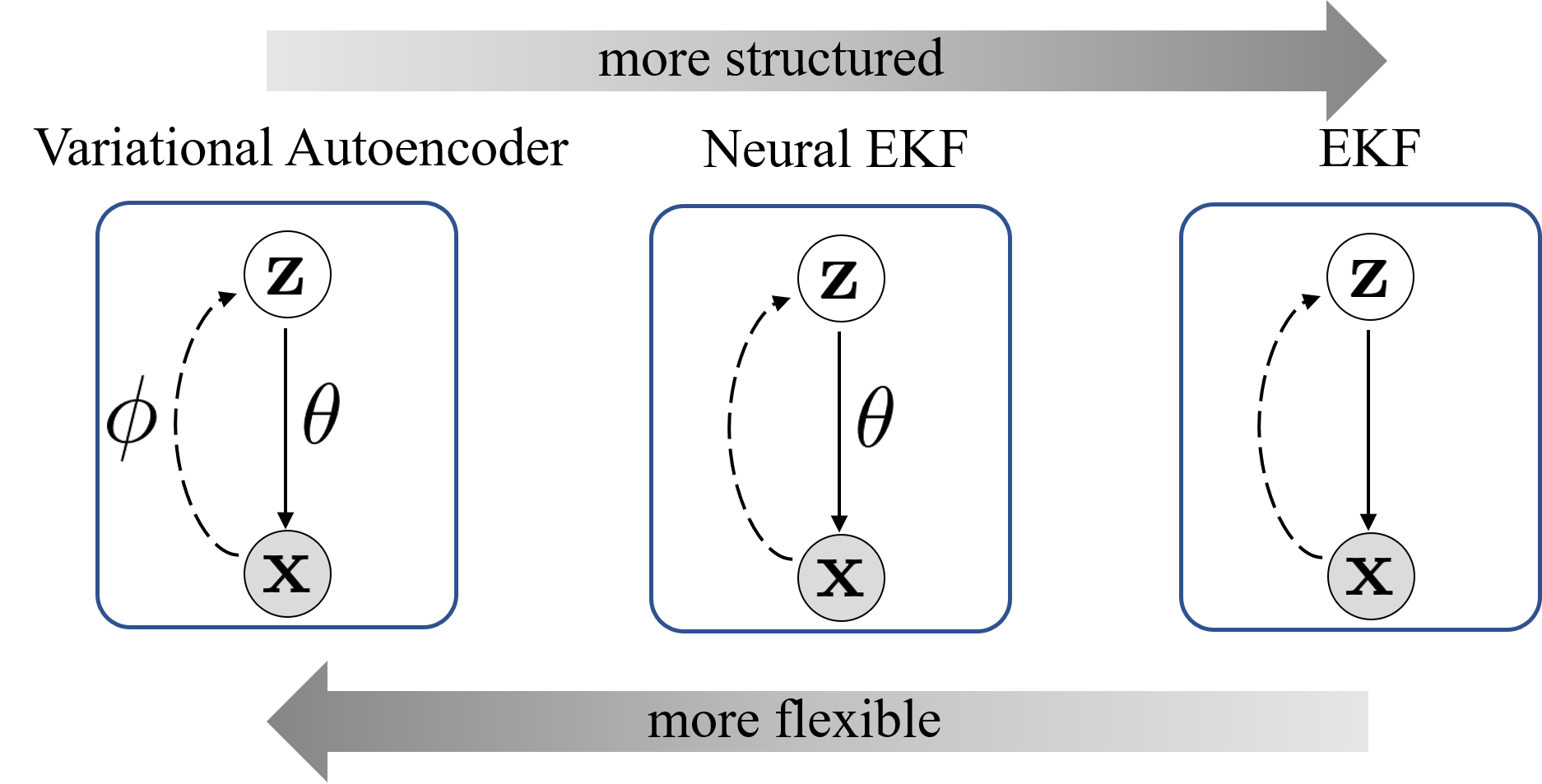}
\caption{Comparison of the variational autoencoder, Neural EKF, and EKF, using a single node as an instance. Solid lines denote the dynamics model. In the VAE and the Neural EKF, the transition model $p_{\theta_\mathbf{t}}(\mathbf{z})$ and observation model  $p_{\theta_\mathbf{o}}(\mathbf{x}|\mathbf{z})$, are parameterized by neural networks with parameters $\theta=\theta_\textbf{t}\bigcup\theta_\textbf{o}$; dashed lines denote the inference model. \textbf{Variational autoencoder}: the inference network $q_\phi(\mathbf{z}|\mathbf{x})$ is configured using a separate set of parameters $\phi$ than the generative model parameters $\theta$. Throughout this paper, the Deep Markov Model (DMM) is adopted as the standard dynamical VAE. \textbf{Neural EKF}: there are no additional parameters $\phi$ involved for the inference model. Instead, as inference is performed via the linearized Kalman filter equations, the inference model implicitly uses $\theta$ that parameterizes the dynamics model. The training process exclusively relies on the parameters $\theta$, thus resulting in a more accurate dynamics model. \textbf{EKF}: there are no learnable parameters involved, and all the governing equations are known, as dictated by the structure of conventional Extended Kalman Filters.}
\label{fig:comparison}
\end{figure}

From the perspective of Kalman filtering, the Neural EKF is a learnable implementation of the EKF, where the transition and observation models can be learned by end-to-end training. 
From the perspective of VAE modeling, the Neural EKF draws from a variational learning methodology with the EKF serving as the inference model, which conducts inference using only dynamics models and does not impose additional hyper-parameters to be learned. We use Figure \ref{fig:comparison} to summarize the comparisons between VAEs, the Neural EKF and the EKF. We validate the framework on three different scenarios to demonstrate its value in learning dynamical systems. Further to verification on simulated data, two experimental case studies on seismic and wind turbine monitoring are presented to validate the capability of the framework to capture the underlying dynamics of complex physical systems. The scheme is shown to perform accurate predictions, which is essential for adoption in downward applications, such as structural health monitoring and decision making.

\section{Preliminaries}

\subsection{Variational Autoencoders and Deep Markov Models}
Variational autoencoders (VAEs) \cite{kingma2013auto} are popularly used for capturing useful latent representations $\textbf{z}$ from data $\textbf{x}$ by reconstructing the VAE input to an output vector. The VAE can be extended to a dynamical version by considering the temporal evolution of the latent variables \cite{girin2020dynamical}. Among various dynamical VAEs, Deep Markov Models (DMMs \cite{krishnan2017structured}) offer a primary example of unsupervised training of a deep state-space model. By chaining an approximate inference model with a generative model and using the variational lower bound maximization approach, DMMs deliver the most direct extension of static VAEs to account for temporal data \cite{girin2020dynamical}. The DMM couples a transition model $p_\theta(\mathbf{z}_t|\mathbf{z}_{t-1})$, describing temporal evolution of the latent states, with an observation model $p_\theta(\mathbf{z}_t|\mathbf{x}_t)$, serving as a decoder and describing the process from latent states to observations, and an inference model $q_\phi(\mathbf{z}_t|\mathbf{x}_{1:T})$, serving as an encoder, which approximates the true posterior distribution $p(\mathbf{z}_t|\mathbf{x}_{1:T})$.

The training of a VAE consists in maximizing a variational lower bound of the data log-likelihood $\log p(\mathbf{x})$. Upon application of the variational principle on the inference model $q_\phi(\mathbf{z}_t|\mathbf{x}_{1:T})$, the evidence lower bound (ELBO) of the data log-likelihood is given as follows:
\begin{equation}\label{ELBO}
\begin{split}
\log p_\theta(\mathbf{x})\geq\mathcal{L}(\theta,\phi;\mathbf{x})&:=\mathbb{E}_{q_\phi(\mathbf{z}|\mathbf{x})}[\log p_{\theta_\mathbf{e}}(\mathbf{x}|\mathbf{z})]\\
&-\text{KL}[q_\phi(\mathbf{z}|\mathbf{x})||p_{\theta_\mathbf{t}}(\mathbf{z}|\mathbf{x})],
\end{split}
\end{equation}
where the KL-divergence is defined as $\text{KL}[q(\mathbf{z})||p(\mathbf{z})] := \int q(\mathbf{z}) \log\frac{q(\mathbf{z})}{p(\mathbf{z})} d\mathbf{z}$.
\begin{figure}[h]
\centering
\includegraphics[width=\linewidth]{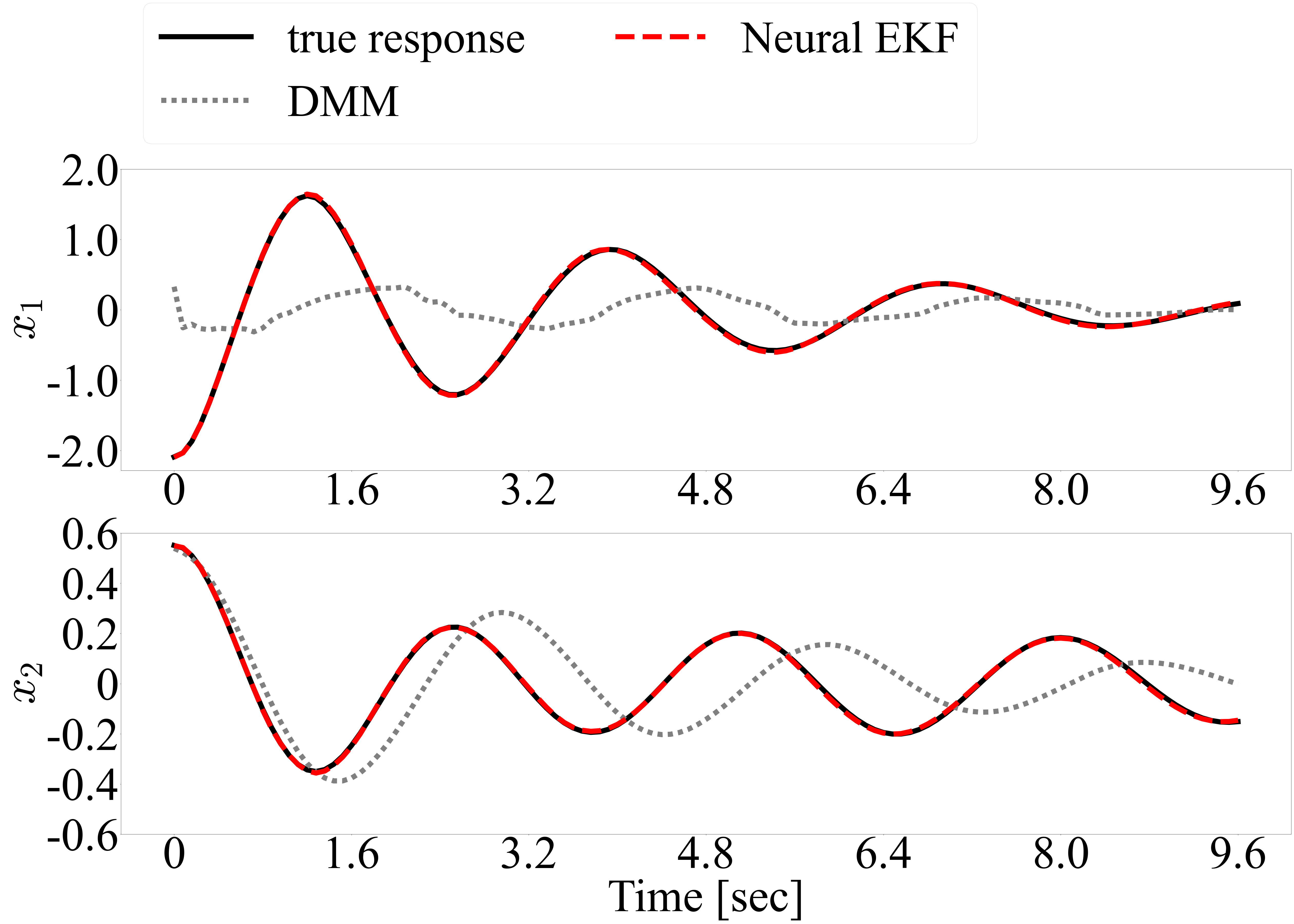}
\caption{Comparison of predictive capabilities of the DMM versus a Neural EKF on a 2-DOF Duffing oscillator system. The displacement response is assumed to serve as the measurements set.}
\label{fig:duffing_comparison}
\end{figure}
Apart from the parameters $\theta$ involved in transition and observation models, the VAE architecture typically employs an additional inference network, which is independently parameterized by the hyper-parameters $\phi$ of the corresponding neural networks. The key idea of VAE is to approximate the posterior distribution with the introduced inference network. However, the objective function ELBO largely depends on the quality of the inference model through the first reconstruction term, which does not involve a transition model, and the influence of the transition models is indirectly relegated to a regularizer for the inference model, reflected in the second KL-divergence term \cite{li2021replay}. The KL-divergence itself is a regularization term and it attempts to only make the single-step dynamics transition similar to the posterior given by the inference network. The above arguments indicate that the quality of VAE reconstructions, as measured by the value of the objective ELBO, strongly depends on the inference model but rather weakly on the transition models. This characteristic of the training objective makes the learned model capable of delivering accurate inference results under availability of observation (monitoring) data, but unsuitable for prediction due to the insufficient training of the transition models.

As shown in Fig. \ref{fig:duffing_comparison}, the result demonstrates the advantages of the Neural EKF over the DMM, which is, as mentioned, a dynamical VAE model that directly extends the static VAE to account for temporal data. The results are demonstrated on a 2-DOF Duffing system, defined by Equation \eqref{eq:duffing} appearing in the Applications section. It is observed that the DMM fails to deliver sufficiently accurate predictions, while the predicted responses, furnished by the Neural EKF, very well fit the true (reference) responses.

\subsection{Extended Kalman Filtering and Smoothing}\label{sec:Kalman}

For nonlinear state estimation and parameter identification in civil and mechanical engineering, the Extended Kalman Filter (EKF) has been a popular choice for weakly nonlinear systems, mainly due to its ease of implementation, robustness and suitability for real-time applications. It assumes a sequence of measurements $\mathbf{x}_{1:T}$ from a monitored dynamical system, which are generated by some latent states $\mathbf{z}_{1:T}$ that are not necessarily directly observed. The transition from a state $\mathbf{z}_{t-1}$ to the next state  $\mathbf{z}_{t}$ is termed as \textit{transition}, and the process from a state ${z}_{t1}$ to its corresponding observation $\mathbf{x}_{t}$ is termed as \textit{observation}. Compared to dynamical VAEs, the EKF assumes that the transition and observation models are given, as described by the following two equations: 
\begin{align}
    \mathbf{z}_t &= f(\mathbf{z}_{t-1}, \mathbf{u}_{t-1}) + w_t,  \; (transition) \label{transition}\\
    \mathbf{x}_t &= g(\mathbf{z}_t) + v_t, \qquad\quad\;\; (observation) \label{observation}
\end{align}
where $f$ and $g$ are two known linear/nonlinear functions governing the transition and observation models, and $w_t\sim\mathcal{N}(0,\mathbf{Q}_t)$, $v_t\sim\mathcal{N}(0,\mathbf{R}_t)$ are Gaussian noise sources with mean zero and covariances $\mathbf{Q}_t$ and $\mathbf{R}_t$, respectively. EKF assumes that the inferred posterior distributions of latent states are based on past and current observations $\mathbf{x}_{1:t}$ and driving force $\textbf{u}_{1:t}$, and follow a Gaussian distribution:
\begin{equation}
    q(\mathbf{z}_t|\mathbf{x}_{1:t},\mathbf{u}_{1:t})=\mathcal{N}(\mu_{t|t},\Sigma_{t|t})\label{eq:EKF_q}
\end{equation}  
Here $\bm{\mu}_{t|t}$ and $\mathbf{\Sigma}_{t|t}$ represent the mean and covariance of the posterior distribution. The Kalman Filter (KF) \cite{10.1115/1.3662552} offers a closed-form of the posterior distributions $q(\mathbf{z}_t|\mathbf{x}_{1:t},\mathbf{u}_{1:t})$ of latent states and provides exact inference for linear systems. The EKF conducts approximate inference, employing a linearization of the nonlinear equations. For EKF, the inference of the posterior distribution is obtained by iteratively executing the following two steps. The first step \textit{predict} is to compute the posterior distribution $q(\mathbf{z}_t|\mathbf{x}_{1:t-1},\mathbf{u}_{1:t-1})=\mathcal{N}(\mu_{t|t-1},\mathbf{\Sigma}_{t|t-1})$, which is simply based on previous observations $\mathbf{x}_{1:t-1}$:
\begin{align}
    \bm{\mu}_{t|t-1}&=f(\bm{\mu}_{t-1|t-1},\mathbf{u}_{t-1}), \label{eq:predict} \\
    \mathbf{\Sigma}_{t|t-1}&=\mathbf{A}_{t-1}\mathbf{\Sigma}_{t-1|t-1}\mathbf{A}_{t-1}^T+\mathbf{Q}_{t-1},
\end{align}
where $\mathbf{A}_{t-1}=\frac{\partial f(\bm{\mu}_{t-1|t-1},\mathbf{u}_{t-1})}{\partial \bm{\mu}_{t-1|t-1}}$ is the Jacobian of $f$ at $\bm{\mu}_{t-1|t-1}$. The second step \textit{update} pertains in updating the posterior distribution from the \textit{predict} step using the current observation $\mathbf{x}_t$:
\begin{align}
    \mathbf{K}_t&=\mathbf{\Sigma}_{t|t-1}\mathbf{C}_t^T(\mathbf{C}_t\mathbf{\Sigma}_{t|t-1}\mathbf{C}_t^T+\mathbf{R}_t)^{-1},\\
    \bm{\mu}_{t|t}&=\bm{\mu}_{t|t-1}+\mathbf{K}_t[\mathbf{x}_t-g(\bm{\mu}_{t|t-1})],\\
    \mathbf{\Sigma}_{t|t}&=(\mathbf{I}-\mathbf{K}_t\mathbf{C}_t)\mathbf{\Sigma}_{t|t-1},\label{eq:update}
\end{align}
where $\mathbf{C}_t=\frac{\partial g(\bm{\mu}_{t|t-1})}{\partial \bm{\mu}_{t|t-1}}$ is the Jacobian of $g$ at $\bm{\mu}_{t|t-1}$. The Jacobian matrices $\mathbf{A}_t$ and $\mathbf{C}_t$ are used as linear approximations of the original nonlinear functions $f$ and $g$, respectively, for conveniently computing  the involved covariances $\mathbf{\Sigma}_{t|t-1}$ and $\mathbf{\Sigma}_{t|t}$. 

The EKF is a recursive filtering method for conducting inference based on $\mathbf{x}_{1:t}$, i.e., the observations until the present time $t$. Since we assume that a sequence of observations $\mathbf{x}_{1:T}$ is available, it is possible and reasonable to further update the inference of posterior distributions with the whole sequence of observations, with respect to a time step within the sequence. This process is termed as \textit{smoothing}. The Rauch-Tung-Striebel smoothing \cite{rauch1965maximum} is a Kalman smoothing algorithm to infer such posterior distributions $q(\mathbf{z}_t|\mathbf{x}_{1:T},\mathbf{u}_{1:T})=\mathcal{N}(\bm{\mu}_{t|T},\mathbf{\Sigma}_{t|T})$ of latent states based on the whole sequence of available observations. The smoothing process can fully make use of available data when computing the posterior distributions and generate more accurate inference results with adequate information, instead of only utilizing information before each time step, which is the case for the Kalman filtering process. Note that, compared to Eq.\eqref{eq:EKF_q}, now the posterior depends on the whole sequence of $\textbf{x}_{1:T}$ and $\textbf{u}_{1:T}$. Once the EKF process is completed, the smoothing algorithm operates in a backward manner, from time $T$ to $1$, to update the posteriors:
\begin{align}
    \mathbf{K}_t^s&=\mathbf{\Sigma}_{t|t}\mathbf{A}_t^T(\mathbf{\Sigma}_{t+1|t})^{-1},\label{eq:smooth1}\\
    \bm{\mu}_{t|T}&=\bm{\mu}_{t|t}+\mathbf{K}_t^s(\bm{\mu}_{t+1|T}-\bm{\mu}_{t+1|t}),\label{eq:smooth2}\\
    \mathbf{\Sigma}_{t|T}&=\mathbf{\Sigma}_{t|t}+\mathbf{K}_t^s(\mathbf{\Sigma}_{t+1|T}-\mathbf{\Sigma}_{t+1|t})(\mathbf{K}_t^s)^T.\label{eq:smooth3}
\end{align}
The intermediate quantities $\bm{\mu}_{t|t}$, $\bm{\mu}_{t+1|t}$, $\mathbf{\Sigma}_{t|t}$ and $\mathbf{\Sigma}_{t+1|t}$ in this smoothing process are already available at Eqs.(\ref{eq:predict})-(\ref{eq:update}), in the EKF stage.

\subsection{Prior and posterior distributions}\label{sec:distribution}
Here we introduce the prior and posterior distributions that will be used for computing the loss function of Neural EKFs in the later section.
The final posterior distribution after application of the EKF algorithm and the smoothing step is given as $q(\mathbf{z}_t|\mathbf{x}_{1:T},\mathbf{u}_{1:T})=\mathcal{N}(\bm{\mu}_{t|T},\mathbf{\Sigma}_{t|T})$, where $\bm{\mu}_{t|T}$ and $\mathbf{\Sigma}_{t|T}$ are given in Eqs.\eqref{eq:smooth2} and \eqref{eq:smooth3}. Furthermore, given
$\mathbf{z}_{t-1}\sim \mathcal{N}(\bm{\mu}_{t-1|T},\mathbf{\Sigma}_{t-1|T})$, we can compute the distribution of each evolved $\textbf{z}_t$:
\begin{equation}\label{eq:trans}
\begin{split}
&\quad\; p(\mathbf{z}_t|\mathbf{z}_{t-1},\mathbf{u}_{t-1})\\
&=\mathcal{N}(f(\bm{\mu}_{t-1|T},\mathbf{u}_{t-1}),\mathbf{A}_{t-1|T}\mathbf{\Sigma}_{t-1|T}\mathbf{A}_{t-1|T}^T+\mathbf{Q}_{t-1}),
\end{split}
\end{equation}
where $\mathbf{A}_{t-1|T}=\frac{\partial f(\bm{\mu}_{t-1|T},\mathbf{u}_{t-1})}{\partial \bm{\mu}_{t-1|T}}$ is the Jacobin of $f$ at $\bm{\mu}_{t-1|T}$. Subsequently, the distribution of observation $p(\mathbf{x}_t|\mathbf{z}_t)$, given $\mathbf{z}_t\sim\mathcal{N}(\bm{\mu}_{t|T},\mathbf{\Sigma}_{t|T})$, can be derived as 
\begin{equation}\label{eq:obs}
p(\mathbf{x}_t|\mathbf{z}_t)=\mathcal{N}(g(\bm{\mu}_{t|T}),\mathbf{C}_{t|T}\Sigma_{t|T}\mathbf{C}_{t|T}^T+\mathbf{R}_t)
\end{equation}
where $\mathbf{C}_{t|T}=\frac{\partial f(\bm{\mu}_{t|T},\mathbf{u}_t)}{\partial \bm{\mu}_{t|T}}$.

\section{Neural Extended Kalman Filters}
A salient limitation of the EKF when applied to learning dynamical systems lies in the requirement for the transition model $f$ and the observation model $g$ to be explicitly defined or at least of known functional format. However, this is not practically feasible when applied to real-world complex systems. In tackling this limitation, the key idea of the proposed Neural EKF is to replace $f$ and $g$ by learnable functions, typically by neural networks. By doing this, the transition and observation models turn to be trainable and efficiently learned by minimizing the defined loss function, making the Neural EKF a flexible tool for learning the dynamics of complex systems. Also, another benefit compared to VAEs, where the inference model is parameterized by neural networks, is that the inference model of the Neural EKF follows a closed-form model. In this section, we will detail the modeling and training of the Neural EKF framework. 

\subsection{Extended Kalman Filters with Learnable Dynamics Models}
In Eqs.(\ref{transition}) and (\ref{observation}), if we replace $f$ and $g$ by learnable transition and observation functions, the framework can be described as:
\begin{align}
    \mathbf{z}_t&=f_{\theta_\textbf{t}}(\mathbf{z}_{t-1},\mathbf{u}_{t-1})+w_t, \; (transition)\\
    \mathbf{x}_t&=g_{\theta_\textbf{o}}(\mathbf{z}_t)+v_t, \qquad\quad\; (observation)
\end{align}
where $f_{\theta_\textbf{t}}$ and $g_{\theta_\textbf{o}}$ are the learnable functions governing the transition and observation models, both parameterized by neural networks with parameters $\theta=\theta_\textbf{t}\bigcup\theta_\textbf{o}$. The process noise sources $w_t$ and observation noise $v_t$ are assumed to follow Gaussian distributions, with respective time-invariant covariances, i.e., $w_t\sim\mathcal{N}(0,\mathbf{Q})$ and $v_t\sim\mathcal{N}(0,\mathbf{R})$ for all time steps $t$. The time-invariant covariances $\mathbf{Q}$ and $\mathbf{R}$ are also set as learnable parameters during the training process. {\color{black} It is noted that here the latent dynamics is formulated as a discrete model. Alternatively, continuous models such as Neural Ordinary Differential Equations \cite{chen2018neural} can also be implemented to model the latent dynamics if one wants to treat the latent dynamics as a continuous model, as explained in Appendix.}


The inference model of the Neural EKF follows the format of the EKF.
This is different from the inference model in VAEs, where $q_\phi$ is a separate inference network parameterized by $\phi$ independent of parameters within $f_{\theta_t}$ and $g_{\theta_o}$. Since the objective ELBO largely depends on the goodness of reconstruction and inference, a separate inference network is thus weakening the training of the transition and observation models.
\subsection{Evidence Lower Bound and Training}\label{sec:ELBO}
With Kalman filters conducting inference, the parameters to be learned are summarized in the vector $\theta$, which includes neural network parameters of both the transition model $f_{\theta_\mathbf{t}}$ and the observation model $g_{\theta_\mathbf{o}}$. In addition, the initial values $\bm{\mu}_{0|0}$ and $\mathbf{\Sigma}_{0|0}$ and covariances $\mathbf{Q}$, $\mathbf{R}$ for respective noises are also parameters to be learned. Similar to the VAE, the training of Neural EKFs is embedded in the variational inference methodology, with the EKF algorithm charged with conducting inference. 
Given any inference model $q_\phi(\mathbf{z}_t|\mathbf{x})$ and Markovian property implied by the dynamics, the ELBO in Eq.\eqref{ELBO} can be expressed in a factorized form ($\mathbf{x}_{1:T}$ and $\mathbf{u}_{1:T}$ are abbreviated as $\mathbf{x}$ and $\mathbf{u}$ in the following formulations for simplicity):
\begin{equation}\label{ELBO_fac}
\begin{split}
&\mathcal{L}(\theta,\phi;\mathbf{x})=\sum_{t=1}^T\Big(\mathbb{E}_{q_\phi(\mathbf{z}_t|\mathbf{x},\mathbf{u})}[\log p_{\theta_\textbf{o}}(\mathbf{x}_t|\mathbf{z}_t)]\\
&-\mathbb{E}_{q_\phi(\mathbf{z}_{t-1}|\mathbf{x},\mathbf{u})}\big[\text{KL}\big(q_\phi(\mathbf{z}_t|\mathbf{x},\mathbf{u})||p_{\theta_\textbf{t}}(\mathbf{z}_t|\mathbf{z}_{t-1},\mathbf{u}_{t-1})\big)\big]\Big),
\end{split}
\end{equation}
where, in the Neural EKF, $q_\phi(\mathbf{z}_t|\mathbf{x},\mathbf{u})$ is actually $q_\theta(\mathbf{z}_t|\mathbf{x},\mathbf{u})$, which alleviates the requirement of additional parameters, further to $\theta$, within the transition and observation models, and thus $\mathcal{L}(\theta,\phi;\mathbf{x})$ reduces to $\mathcal{L}(\theta;\mathbf{x})$. Since the posterior distributions $q_\theta(\mathbf{z}_t|\mathbf{x},\mathbf{u})$ can be computed in closed form by EKF, the distributions in Eq.\eqref{ELBO_fac} can thus be computed explicitly given $\mathbf{z}_t\sim\mathcal{N}(\bm{\mu}_{t|T},\mathbf{\Sigma}_{t|T})$, as shown in Section \textit{Prior and posterior distributions}. Thus,  the ELBO Eq.\eqref{ELBO_fac} can be computed in a surrogate way as:
\begin{equation}\label{ELBO_EKF}
\begin{split}
&\mathcal{L}(\theta;\mathbf{x})=\sum_{t=1}^T\Big(\log p_{\theta_\textbf{o}}(\mathbf{x}_t|\mathbf{z}_t)\\
&-\text{KL}\big(q_\theta(\mathbf{z}_t|\mathbf{x},\mathbf{u})||p_{\theta_\textbf{t}}(\mathbf{z}_t|\mathbf{z}_{t-1},\mathbf{u}_{t-1})\Big), \\
&\text{given } \mathbf{z}_t\sim q_\theta(\mathbf{z}_t|\mathbf{x},\mathbf{u}).
\end{split}
\end{equation}
The above involved distributions are based on the prior $\mathbf{z}_t\sim\mathcal{N}(\bm{\mu}_{t|T},\mathbf{\Sigma}_{t|T})$. The log-probability and KL-divergence can be computed analytically when all the involved distributions are Gaussian. The log-probability of a Gaussian distribution $p(\mathbf{x})=\mathcal{N}(\bm{\mu},\mathbf{\Sigma})$ has an explicit format as:
\begin{equation}\label{eq:likelihood}
\begin{split}
\log p(\mathbf{x})&=-\frac{1}{2}\big[\log|\mathbf{\Sigma}|+(\mathbf{x}-\bm{\mu})^T\mathbf{\Sigma}^{-1}(\mathbf{x}-\bm{\mu})\\
&+d_x\log(2\pi)\big]
\end{split}
\end{equation}
where $d_x$ is the dimension of $\mathbf{x}$. As stated in Eq.\eqref{eq:obs}, since $\mathbf{z}_t\sim q_\theta(\mathbf{z}_t|\mathbf{x},\mathbf{u})=\mathcal{N}(\bm{\mu}_{t|T},\mathbf{\Sigma}_{t|T})$ and replace $g$ by $g_{\theta_\textbf{o}}$, we have $p(\mathbf{x}_t|\mathbf{z}_t)=\mathcal{N}(g_{\theta_\textbf{o}}(\bm{\mu}_{t|T}),\mathbf{C}_{t|T}\Sigma_{t|T}\mathbf{C}_{t|T}^T+\mathbf{R})$ . Using  Eq.\eqref{eq:likelihood}, the log-likelihood term $\log p_{\theta_\textbf{o}}(\mathbf{x}_t|\mathbf{z}_t)$ in Eq.\eqref{ELBO_EKF} can be computed approximately as:
\begin{equation}
\begin{split}
&\log p_{\theta_\textbf{o}}(\mathbf{x}_t|\mathbf{z}_t) \text{ given } \mathbf{z}_t\sim q_\theta(\mathbf{z}_t|\mathbf{x},\mathbf{u}) \\
=&-\frac{1}{2}\big[\log|\mathbf{C}_{t|T}\mathbf{\Sigma}_{t|T}\mathbf{C}_{t|T}^T+\mathbf{R}|+d_x\log(2\pi)\\
&+(\mathbf{x}_t-g_{\theta_\textbf{o}}(\bm{\mu}_{t|T}))^T(\mathbf{C}_{t|T}\mathbf{\Sigma}_{t|T}\mathbf{C}_{t|T}^T+\mathbf{R})^{-1}\\
&(\mathbf{x}_t-g_{\theta_\textbf{o}}(\bm{\mu}_{t|T}))\big]\label{eq:posterior1}
\end{split}
\end{equation}
Similarly, the KL-divergence term $\text{KL}(q(\mathbf{z})||p(\mathbf{z}))$, when  $p(\mathbf{z})=\mathcal{N}(\bm{\mu}_p,\mathbf{\Sigma}_p)$ and $q(\mathbf{z})=\mathcal{N}(\bm{\mu}_q,\mathbf{\Sigma}_q)$ are both Gaussian distributions, comprises an analytical form as: 
\begin{equation}\label{eq:KL}
\begin{split}
\text{KL}(q(\mathbf{z})||p(\mathbf{z}))&=\frac{1}{2}\Big[\log\frac{|\mathbf{\Sigma}_{p}|}{|\mathbf{\Sigma}_{q}|}-d_z+\text{Tr}(\mathbf{\Sigma}_{p}^{-1}\mathbf{\Sigma}_{q})\\
&+(\bm{\mu}_{p}-\bm{\mu}_{q})^T\mathbf{\Sigma}_{p}^{-1}(\bm{\mu}_{p}-\bm{\mu}_{q})\Big],
\end{split}
\end{equation}
where $d_z$ is the dimension of $\mathbf{z}$. As stated in Eq.\eqref{eq:trans}, and replace $f$ by $f_{\theta_\textbf{t}}$,
\begin{equation}
\begin{split}
&p_{\theta_\textbf{t}}(\mathbf{z}_t|\mathbf{z}_{t-1},\mathbf{u}_{t-1})\\
=&\mathcal{N}(f_{\theta_\textbf{t}}(\bm{\mu}_{t-1|T},\mathbf{u}_{t-1}),\mathbf{A}_{t-1|T}\mathbf{\Sigma}_{t-1|T}\mathbf{A}_{t-1|T}^T+\mathbf{Q})
\end{split}
\end{equation}
since we have $\mathbf{z}_{t-1}\sim q_\theta(\mathbf{z}_{t-1}|\mathbf{x},\mathbf{u})=\mathcal{N}(\bm{\mu}_{t-1|T},\mathbf{\Sigma}_{t-1|T})$ and  $\mathbf{z}_{t}\sim q_\theta(\mathbf{z}_{t}|\mathbf{x},\mathbf{u})=\mathcal{N}(\bm{\mu}_{t|T},\mathbf{\Sigma}_{t|T})$. Substitute these mean and covariance into Eq.\eqref{eq:KL}, the KL-divergence term in Eq.\eqref{ELBO_EKF} can be computed as:
\begin{equation}
\begin{split}
&\text{KL}(q_\theta(\mathbf{z}_t|\mathbf{x},\mathbf{u})||p_{\theta_\textbf{t}}(\mathbf{z}_t|\mathbf{z}_{t-1},\mathbf{u}_{t-1}))\\
&\text{given } \mathbf{z}_{t-1}\sim q_\theta(\mathbf{z}_{t-1}|\mathbf{x},\mathbf{u})=\mathcal{N}(\bm{\mu}_{t-1|T},\mathbf{\Sigma}_{t-1|T})\\
=&\frac{1}{2}\Big[\log\frac{|\mathbf{A}_{t-1|T}\mathbf{\Sigma}_{t-1|T}\mathbf{A}_{t-1|T}^T+\mathbf{Q}|}{|\mathbf{\Sigma}_{t|T}|}-d_z\\
+&\text{Tr}((\mathbf{A}_{t-1|T}\mathbf{\Sigma}_{t-1|T}\mathbf{A}_{t-1|T}^T+\mathbf{Q})^{-1}\mathbf{\Sigma}_{t|T})\\
+&(f_{\theta_\textbf{t}}(\bm{\mu}_{t-1|T})-\bm{\mu}_{t|T})^T(\mathbf{A}_{t-1|T}\Sigma_{t-1|T}\mathbf{A}_{t-1|T}^T+\mathbf{Q})^{-1}\\
&(f_{\theta_\textbf{t}}(\bm{\mu}_{t-1|T})-\bm{\mu}_{t|T})\Big]
\end{split}
\end{equation}

Thus the final objective ELBO enforced by EKF and smoothing formulas can be written as:
\begin{equation}\label{ELBO_analytic}
\begin{split}
&\mathcal{L}(\theta;\mathbf{x})
= -\frac{1}{2}\sum_{t=1}^{T}\Big[\log|\mathbf{C}_{t|T}\mathbf{\Sigma}_{t|T}\mathbf{C}_{t|T}^T+\mathbf{R}|\\
&+(\mathbf{x}_t-g_{\theta_\textbf{o}}(\bm{\mu}_{t|T}))^T(\mathbf{C}_{t|T}\mathbf{\Sigma}_{t|T}\mathbf{C}_{t|T}^T)^{-1}(\mathbf{x}_t-g_{\theta_\textbf{o}}(\bm{\mu}_{t|T}))\\ &+d_x\log(2\pi)
+\log\frac{|\mathbf{A}_{t-1|T}\mathbf{\Sigma}_{t-1|T}\mathbf{A}_{t-1|T}^T+\mathbf{Q}|}{|\mathbf{\Sigma}_{t|T}|}-d_z\\
&+\text{Tr}((\mathbf{A}_{t-1|T}\mathbf{\Sigma}_{t-1|T}\mathbf{A}_{t-1|T}^T+\mathbf{Q})^{-1}\mathbf{\Sigma}_{t|T})\\&
+(f_{\theta_\textbf{t}}(\bm{\mu}_{t-1|T})-\bm{\mu}_{t|T})^T(\mathbf{A}_{t-1|T}\Sigma_{t-1|T}\mathbf{A}_{t-1|T}^T\\
&+\mathbf{Q})^{-1}(f_{\theta_\textbf{t}}(\bm{\mu}_{t-1|T})-\bm{\mu}_{t|T})\Big].
\end{split}
\end{equation}

Typically, the variational objective function for the dynamical VAE framework focuses on the reconstruction loss, which largely depends on the inference model (encoder) and observation model (decoder), whereas the transition model plays a minor role in the variational objective and works as an intermediate process for training. This often results in an accurate inference model and a less meaningful transition model, which is not applicable for prediction because the transition model cannot reflect the true underlying dynamics. Therefore, it is necessary that the objective function also takes the accuracy of the transition model into consideration to ensure its closeness to the true latent dynamic process. To address this issue, we adopt the overshooting method proposed in \cite{li2021replay}, which is termed as \textit{replay overshooting}. The key point lies in simply introducing the prediction loss of the generative model into the objective function. After obtaining a sequence of means and covariances $\{(\mu_{t|T},\Sigma_{t|T})\}$, the initial value $(\bar{\mu}_0,\bar{\Sigma}_0)=(\mu_{0|T},\Sigma_{0|T})$ will be further used for prediction process. Then similarly as the prediction step in EKF, the predicted values are obtained via the generative model as:
\begin{align}
    \bar{\bm{\mu}}_t&=f(\bar{\bm{\mu}}_{t-1},\mathbf{u}_t),\label{eq:ro1}\\
    \bar{\mathbf{\Sigma}}_t&=\mathbf{A}_{t-1}\bar{\mathbf{\Sigma}}_{t-1}\mathbf{A}_{t-1}^T+\mathbf{Q},\label{eq:ro2}
\end{align}
and we receive another set of distributions $\bar{q}(\mathbf{z}_t)=\mathcal{N}(\bar{\mu}_t,\bar{\Sigma}_t)$ from the generative model. The final reconstruction loss is composed of the both sets of posterior distributions, i.e., the posterior distributions obtained from the Kalman inference model in Eq.\eqref{eq:posterior1}, as well as those obtained from the transition model, weighted by $\alpha$ (in this paper, we set $\alpha = 0.5$). These form the objective function combined with the KL loss, expressed as:
\begin{equation}
\begin{split}
\mathcal{L}(\theta;\textbf{x})&=\sum_{t=1}^T\Big(\alpha\mathbb{E}_{q_\theta(\mathbf{z}_t)}[\log p_\theta(\mathbf{x}_t|\mathbf{z}_t)]\\
&+(1-\alpha)\mathbb{E}_{\bar{q}_\theta(\mathbf{z}_t)}[\log p_\theta(\mathbf{x}_t|\mathbf{z}_t)]\\
&-\mathbb{E}_{q_\theta(\mathbf{z}_{t-1})}[\text{KL}(q(\mathbf{z}_t)||p(\mathbf{z}_t|\mathbf{z}_{t-1},\mathbf{u}_t))]\Big),
\end{split}
\end{equation}
where the first and third term are the same as in \eqref{ELBO_analytic}, while the second term is for overshooting.

The pipeline of the Neural EKF is summarized in Algorithm \ref{algorithm}.
\begin{algorithm}[H]
\caption{Neural Extended Kalman Filters}\label{algorithm}
\begin{algorithmic}
\State Initialize parameters $\theta,\bm{\mu}_{0|0},\mathbf{\Sigma}_{0|0},\mathbf{Q},\mathbf{R}$
\While{$\theta$ not converged}
\For{batch $b=1,...,B$}
\State $(\bm{\mu}_{t|t},\mathbf{\Sigma}_{t|t})=\text{EKF}(\theta,\mathbf{x}_{1:T},\mathbf{Q},\mathbf{R})$\\
\quad (Extended Kalman Filtering using Eqs.\eqref{eq:predict}-\eqref{eq:update})
\State $(\bm{\mu}_{t|T},\mathbf{\Sigma}_{t|T})=\text{EKS}(\theta,\mathbf{x}_{1:T},\mathbf{Q},\mathbf{R})$\\
\quad (Extended Kalman Smoothing using Eqs.\eqref{eq:smooth1}-\eqref{eq:smooth3})
\State $(\bar{\bm{\mu}}_t,\bar{\mathbf{\Sigma}}_t)=\text{RO}(\theta,\bm{\mu}_{0|0},\mathbf{\Sigma}_{0|0},\mathbf{Q},\mathbf{R})$\\
\quad (Replay overshooting using Eqs.\eqref{eq:ro1} and \eqref{eq:ro2})
\State Compute the objective value $ \mathcal{L}(\theta;\mathbf{x})$
\State Update $\theta,\bm{\mu}_{0|0},\mathbf{\Sigma}_{0|0},\mathbf{Q},\mathbf{R}$ with stochastic
\State gradient ascent on $\mathcal{L}$
\EndFor
\EndWhile
\end{algorithmic}
\end{algorithm}
\subsection{Neural EKF for Response Predictions}
Once a Neural EKF model is trained, the learned function $f_{\theta_\textbf{t}}$ and $g_{\theta_\textbf{o}}$ is available and can be used for response prediction. The prediction is performed using     $\mathbf{z}_t=f_{\theta_\textbf{t}}(\mathbf{z}_{t-1},\mathbf{u}_{t-1})$ and $\mathbf{x}_t=g_{\theta_\textbf{o}}(\mathbf{z}_t)$, given a new initial condition $\textbf{z}_0$ and corresponding driving forces $\textbf{u}_{1:T}$. 
Note that a new initial condition can be either specified by the user, or inferred from actual data via use of the inference model of the Neural EKF. 

The Neural EKF is capable of generating predictions in real-time to detect structural anomalies and changes as they occur for structural health monitoring. On the other hand, it requires offline training on a characteristic batch of operational data, prior to it being applied in prediction mode. However, in intervals where loads significantly increase with respect to the training set, possibly triggering more severe nonlinearity, or if the system experiences damage, the model must be updated offline using newly acquired data for training, in order to adapt to the changed dynamics. This is a trade-off that is often present in real-time monitoring systems, where the need for accurate predictions must be balanced with the need for timely updates.


\section{Application Case Studies}
We verify and validate the proposed Neural EKF framework on both simulated and real-world datasets. The numerical simulation aims to demonstrate the predictive capabilities of the learned model via Neural EKF, as compared against the commonly adopted option of variational autoencoders. 

The experiments involving real-world monitoring datasets aim to further demonstrate the Neural EKF's ability to learn the dynamics when the system is considerably complex. In this case, physics-based modeling can be challenging due to the unknown model structure, or computationally unaffordable. In the employed seismic monitoring case study, we validate the effectiveness of the Neural EKF in predicting seismic responses on the basis of available ground motion inputs (earthquakes). Furthermore, in the experimentally tested wind turbine case study, we perform a thorough analysis on dynamic response prediction, damage detection and structural health monitoring based on the learned Neural EKF model.

{\color{black} The transition function $f_{\theta_\textbf{t}}$ and observation functions $g_{\theta_\textbf{o}}$ are modeled as multilayer perceptrons consisting of three hidden layers. Each hidden layer comprises 64 nodes for the Duffing and seismic response examples, and 128 nodes for the wind turbine example. The dimension of the latent states $\mathbf{z}_t$ is assumed to correspond to twice the DOFs of the modeled system, which is an assumption commonly adopted in conventional state-space modeling.} The dimension of latent states is a predefined hyperparameter and can be fine-tuned if needed. This is different from conventional Kalman filters, where a fixed state equation, and thus a corresponding dimension, has to be assumed. In this case, issues may arise in terms of the observability of the system's states, in case the observed variables are not carefully chosen, and particularly if they are lower than the number of predefined latent states. However, in our proposed method, the latent state mainly serves as an intermediate quantity to optimally reconstruct and predict the observed quantities. Similar to the width and depth of neural networks, a larger latent state dimension usually benefits the training result but may lead to overfitting. In our examples, we opted for a latent state dimension that equals twice the number of DOFs of the modeled system, as this is a natural choice. The data and codes used in this paper are publicly available on GitHub at https://github.com/liouvill/NeuralEKF.

\subsection{Duffing Oscillator System}
\begin{figure}[h]
\centering
\includegraphics[width=0.95\linewidth]{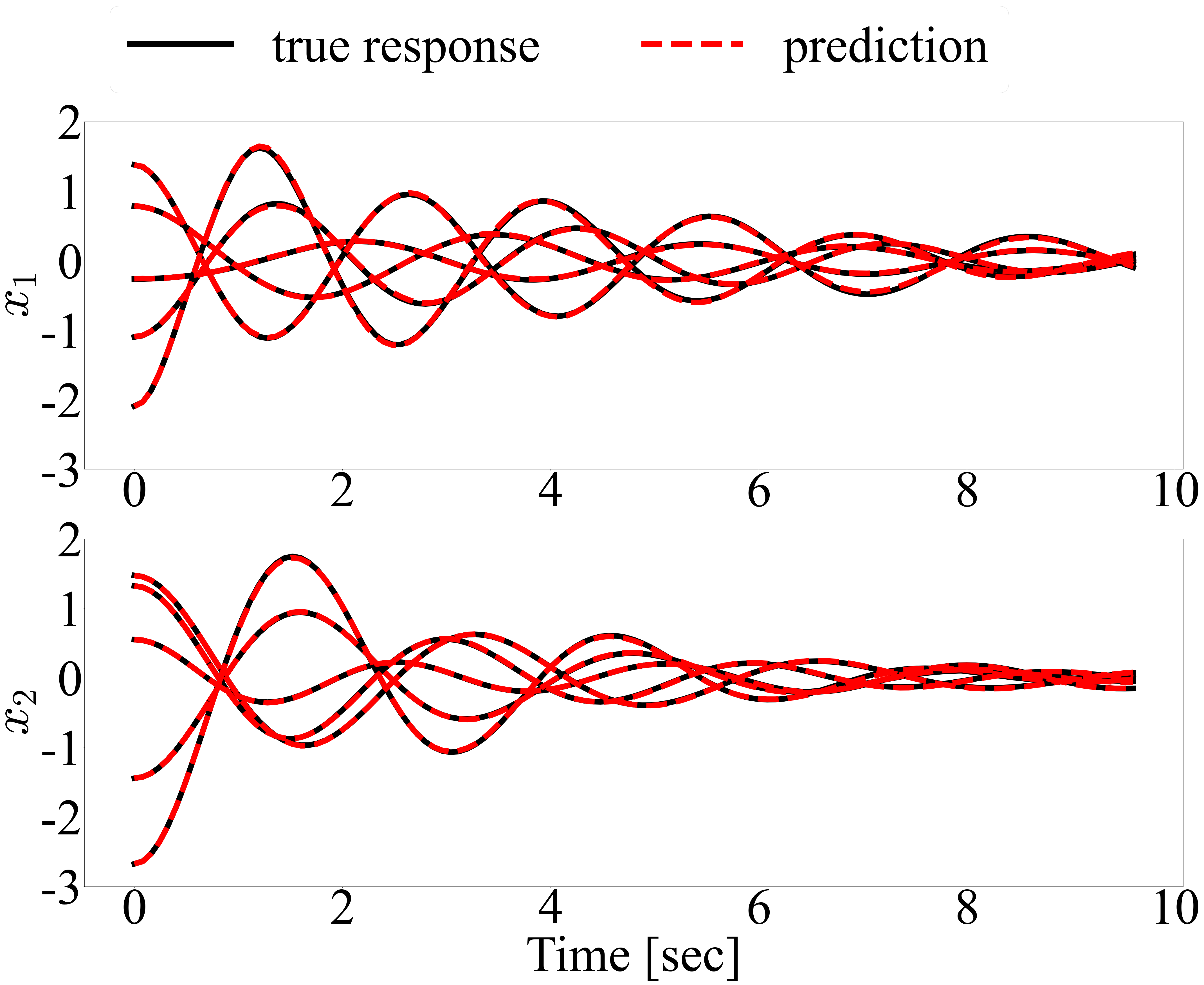}
\caption{Predicted results for the simulated 2DOF Duffing system.}
\label{duffing}
\end{figure}
We first demonstrate the performance of the Neural EKF framework for predicting the dynamic response through a numerical example. In this example, we consider a 2 degrees-of-freedom (DOF) nonlinear duffing oscillator subjected to random excitations. The data used for this case study are simulated by the following differential equation:
\begin{equation}\label{eq:duffing}
    \begin{bmatrix}
    \dot{\mathbf{x}}\\
    \ddot{\mathbf{x}}
    \end{bmatrix}
    = \begin{bmatrix}
    \mathbf{0} & \mathbf{I}\\
    -\mathbf{M}^{-1}\mathbf{K} &   -\mathbf{M}^{-1}\mathbf{C}
    \end{bmatrix}\begin{bmatrix} 
    \mathbf{x}\\
    \dot{\mathbf{x}}
    \end{bmatrix} + \begin{bmatrix}
    0\\0\\
    -k_nx_1^3\\
    0
    \end{bmatrix},
\end{equation}
where $\mathbf{x}=\begin{bmatrix} x_1 \\ x_2 \end{bmatrix}$, the mass matrix $\mathbf{M}=\begin{bmatrix} 1 & 0 \\ 0 & 1 \end{bmatrix}$, the stiffness matrix $\mathbf{K}=\begin{bmatrix} 4 & -0.5 \\ -0.5 & 4 \end{bmatrix}$, the damping matrix $\mathbf{C}=\begin{bmatrix} 0.5 & 0 \\ 0 & 0.5 \end{bmatrix}$ and the coefficient of the cubic stiffness term $k_n=1$. Here the displacements of both DOFs $x_1$ and $x_2$ are assumed available as measurements. It is noted that this parameter setting falls into the scenario where the Duffing oscillator presents a single stable equilibrium point at the origin. We consciously make this choice, as we here adopt this as a nonlinear example, in order to demonstrate the capability of the Neural EKF to deal with nonlinearity, but not rare instabilities or chaotic behavior, which would require a more complex treatment.

For this numerical case study, a number of 1000 random trajectories are generated with both transition and observation noise variances equal to 0.001 for training by the simulated 2-DOF nonlinear Duffing system shown in Eq.(\ref{eq:duffing}). Another 5 randomly generated trajectories are used as an unknown dataset for testing the prediction capability of the Neural EKF framework. The prediction results for the free vibration case are shown in Figure \ref{duffing}. As indicated by the results, the Neural EKF is capable of producing satisfactory predictions that reasonably match the reference well for this simulated nonlinear system. As comparison, which is already shown in Fig \ref{fig:duffing_comparison}, it is observed that the DMM, which is a pure VAE-based method, fails to capture the dynamics and the predictions are much less accurate.

\begin{table}[h]
\caption{Comparison with noise-free system responses under different noise levels: $\text{RMSE}_1$ is for $x_1$ and $\text{RMSE}_2$ is for $x_2$.} 
\centering 
\begin{tabular}{c cc} 
\toprule 
Noise variance &$\text{RMSE}_1$ &$\text{RMSE}_2$ \\
\midrule 
0.001 & 0.04865 & 0.01691 \\  
0.01 & 0.03770 & 0.01331 \\  
0.1 & 0.05192 & 0.01709 \\  
\bottomrule 
\end{tabular}
\label{tab:noise}
\end{table}

\begin{figure*}[h]
\begin{subfigure}{.33\linewidth}
  \centering
  \includegraphics[width=\linewidth]{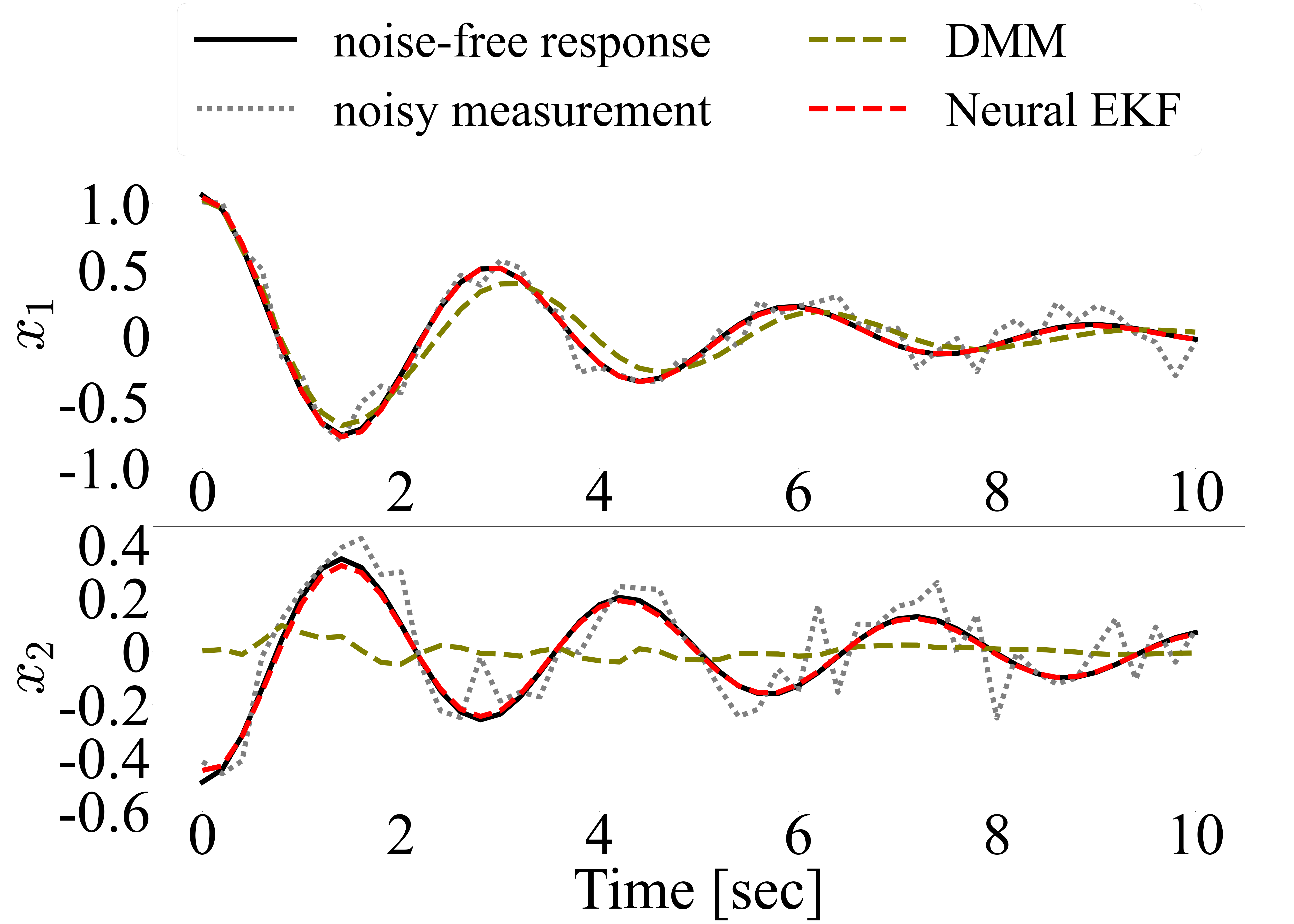}
  \caption{$\text{noise variance}=0.1$}
  \label{fig:noise_0.1}
\end{subfigure}
\begin{subfigure}{.33\linewidth}
  \centering
  \includegraphics[width=\linewidth]{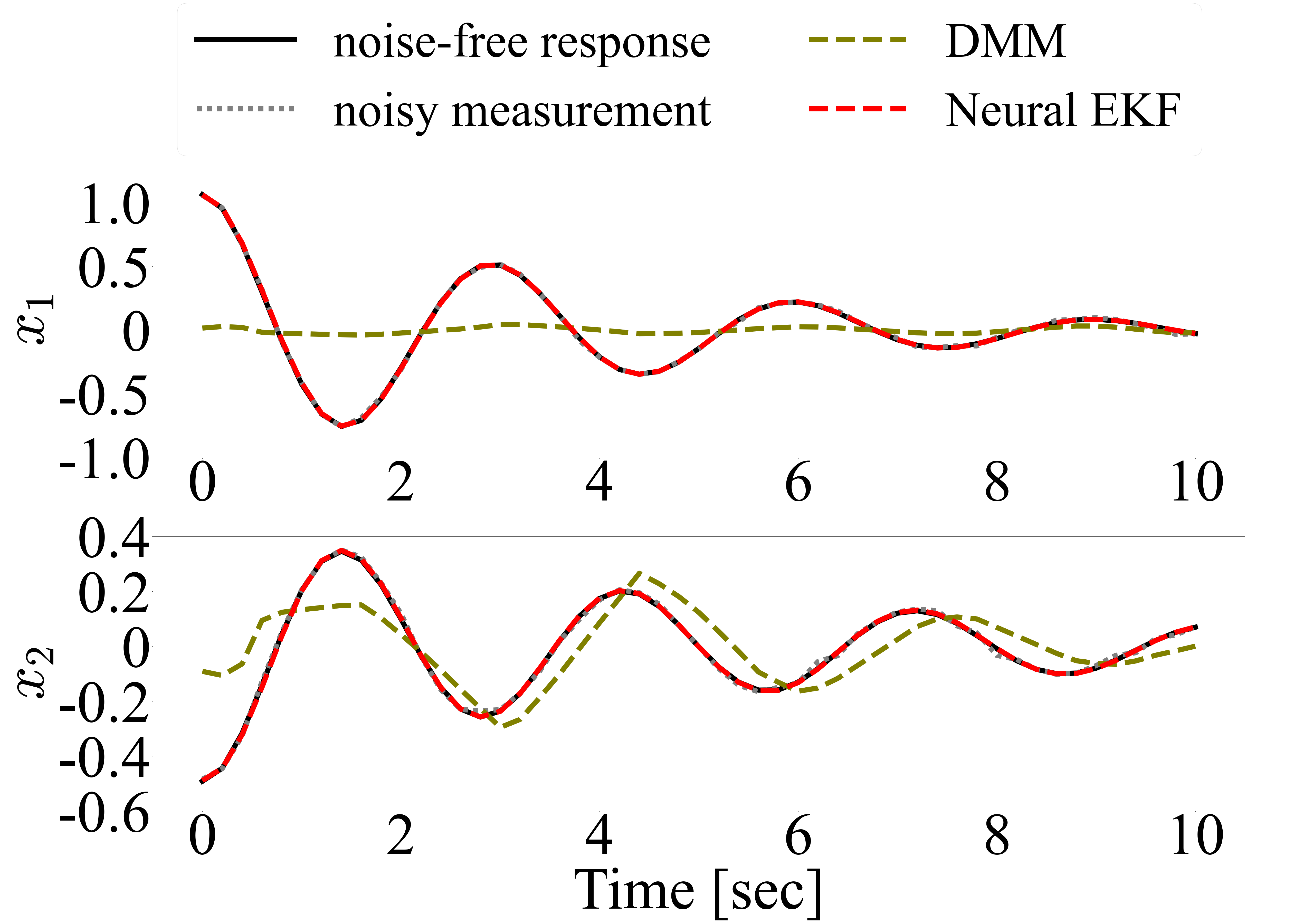}
  \caption{$\text{noise variance}=0.01$}
  \label{fig:noise_0.01}
\end{subfigure}
\begin{subfigure}{.33\linewidth}
  \centering
  \includegraphics[width=\linewidth]{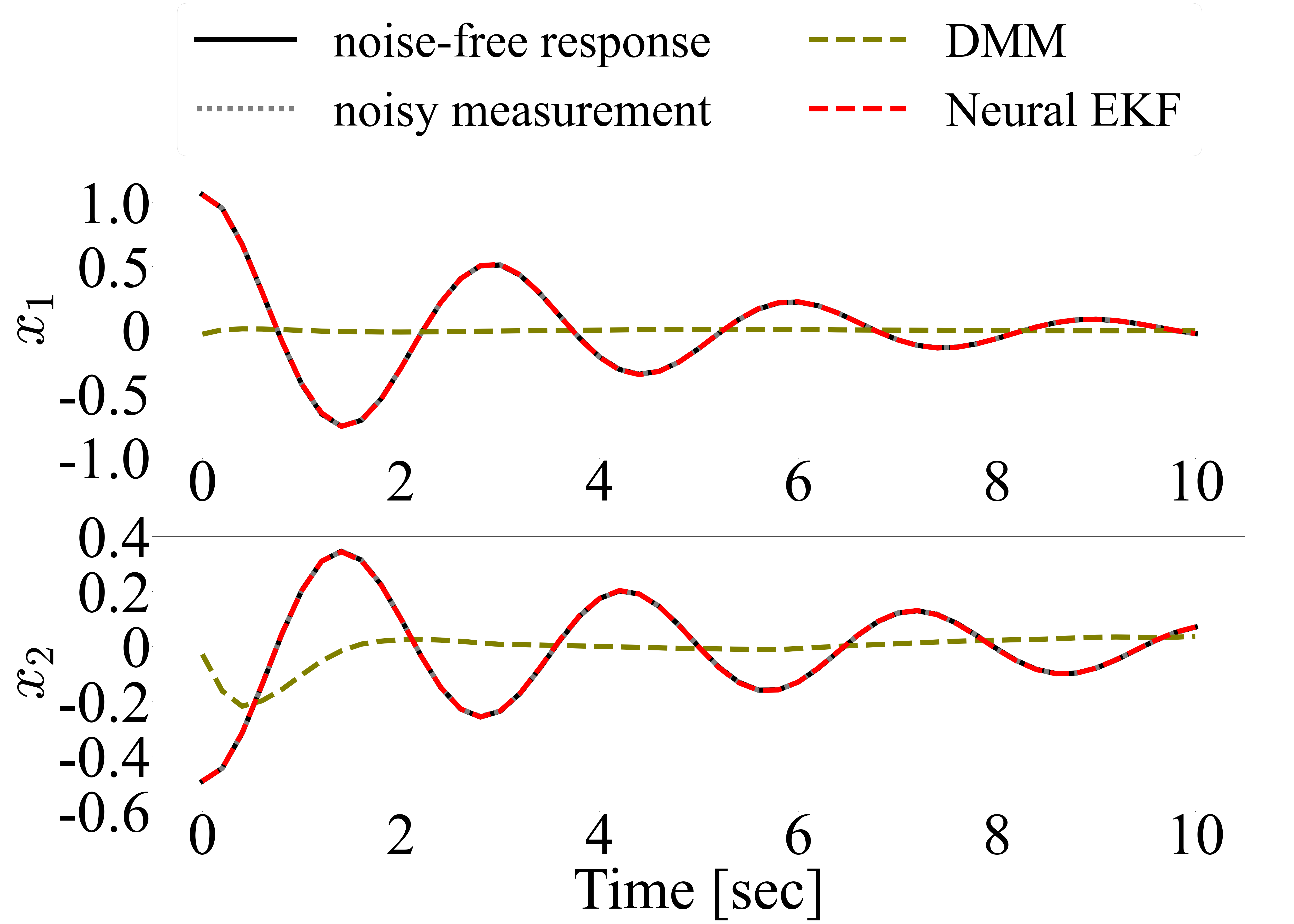}
  \caption{$\text{noise variance}=0.001$}
  \label{fig:noise_0.001}
\end{subfigure}
\caption{Comparison of predicted responses versus noise-free responses. The noisy measurements are used as training data.}
\label{fig:noisy_comparison}
\end{figure*}

{\color{black} To further demonstrate the robustness of the proposed model, we have generated artificial observations (measured data) with different levels of noise contamination. The data is generated by Eq.\eqref{eq:duffing} using the same initial values across three cases, but with different levels of additive Gaussian observation noise. More specifically, we generate three different simulated cases corresponding to an observation noise variance of 0.001, 0.01 and 0.1, respectively. The ``process noise" is set to be zero, when generating artificial observation data, since integration of a larger transition process noise would cause the solution of the ODE to diverge due to nonlinearity of the Duffing system. The root mean squared error between the predicted and true (noise-free) system response is reported in Table \ref{tab:noise}. Our results indicate that the difference between the predicted and noise-free system response, for training and prediction under assumed contaminated measurements, are similar across different noise covariance levels. The proposed method thus proves particularly robust to noise contamination, as revealed by the results shown in Figure \ref{fig:noisy_comparison} for the case with different observation noise variances. As comparison, the DMM-predicted results are plotted in dark green, which deviate significantly from the ground-truth and fail to provide accurate and reliable predictions.
}

\subsection{Seismic Response Data}
\begin{figure}[h]
\centering
\includegraphics[width=1.0\linewidth]{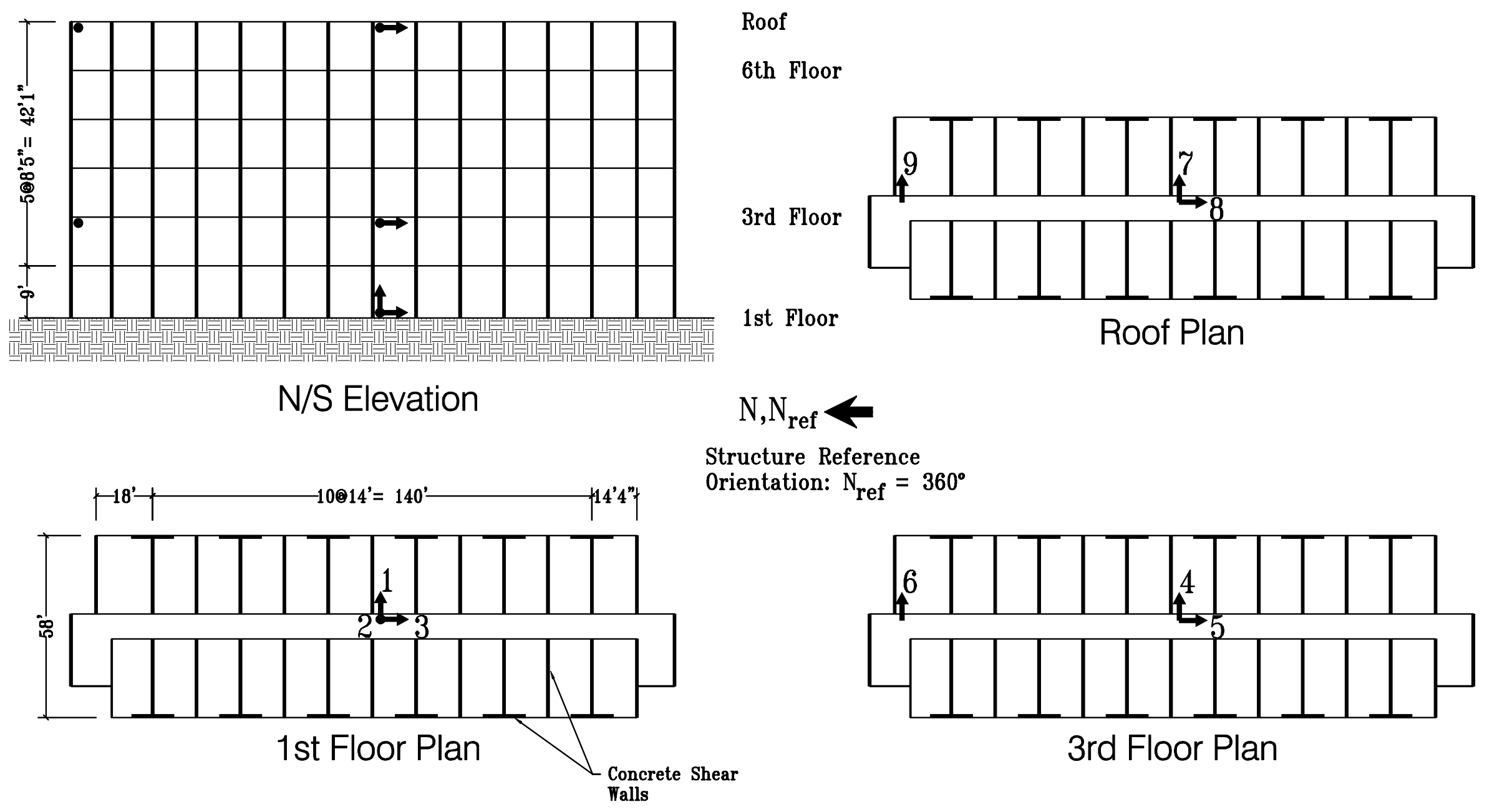}
\caption{Sensor placement of the 6-story hotel building in San Bernardino, California. 
}
\label{building}
\end{figure}
The framework is further investigated using real-world seismic data obtained from the Center for Engineering Strong Motion Data (CESMD) \cite{haddadi2008center} for a 6-story hotel building in San Bernardino, California. It is a mid-rise concrete building installed with a total of nine accelerometers on the first, third and roof floors in different directions. The sensor placement is shown in Figure \ref{building}, where three accelerometers are mounted on each of the ground floor, third floor, and roof. The sensors have recorded multiple seismic events from 1987 to 2021. 17 corresponding structural response and ground motion datasets are used to train the Neural EKF with one dataset reserved for evaluating prediction performance, which corresponds to the San Bernardino earthquake in 2009. The data we used are extracted from sensors 1, 4 and 7, which lie along the same direction. The data from sensor 1 are used as the input ground motion, while the data from sensors 4 and 7 are used as output responses. Due to the inconsistency of the sample frequencies, we first pre-process the data using resampling techniques. The signals are uniformly resampled at 50 Hz (the downloaded data was already pre-processed by the website with noise filtering, baseline and sensor offset corrections).

\begin{figure}
\centering
\includegraphics[width=1.0\linewidth]{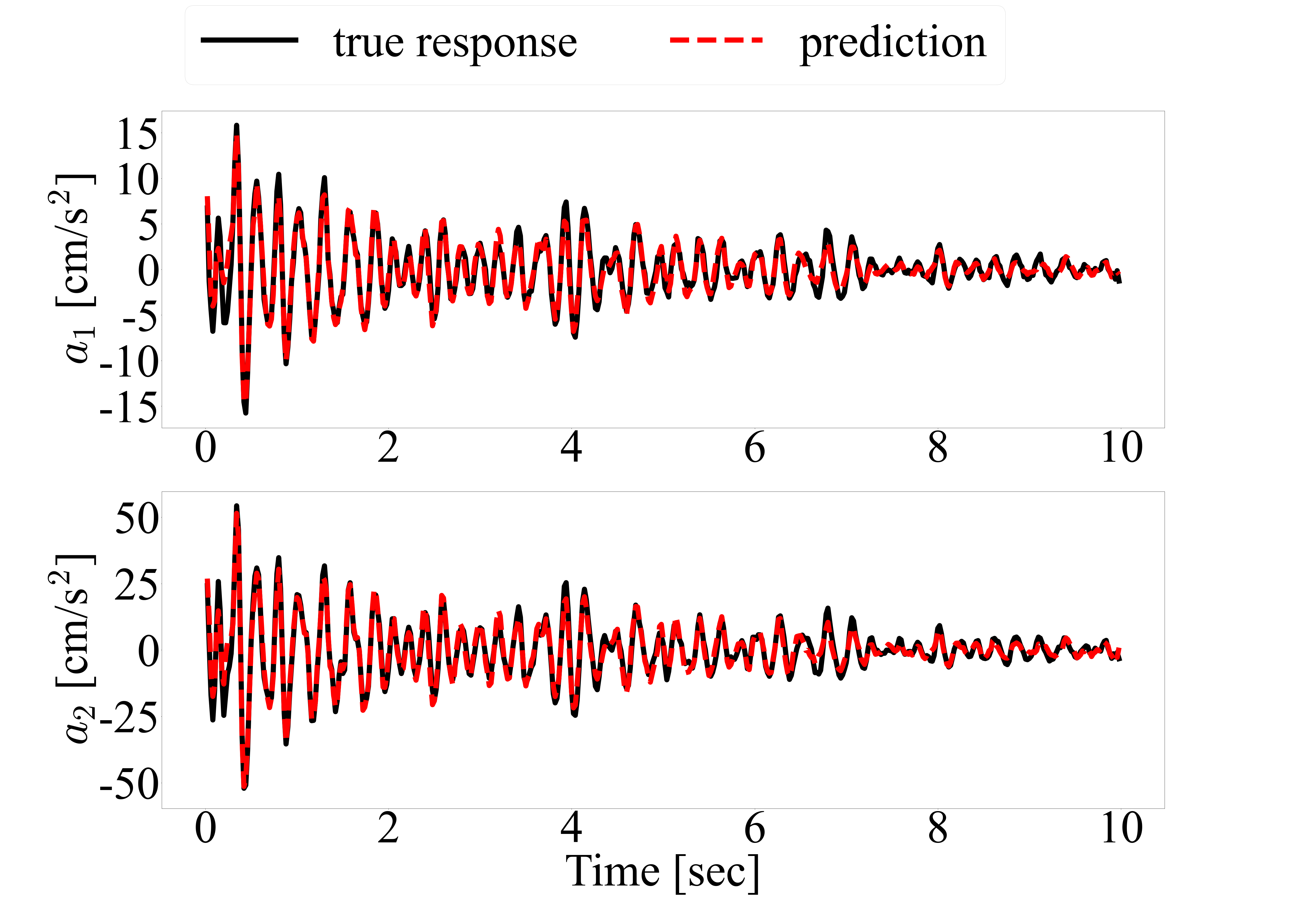}
\caption{Relative accelerations of the building under seismic excitations.}
\label{seismic}
\end{figure}

After training the Neural EKF model with a pre-processed training dataset, the predictions on another independent test dataset are obtained by simply feeding the new seismic inputs into the trained model. The predictions and ground truth values are compared in Figure \ref{seismic}. A strong agreement between prediction responses and ground truth responses can be observed. While the data are sampled at different frequencies and earthquakes vary widely in magnitudes, directions, and frequencies, the Neural EKF essentially learns dynamics of the building, thus generating highly consistent predictions.
\begin{figure}[b]
\centering
\includegraphics[width=0.9\linewidth]{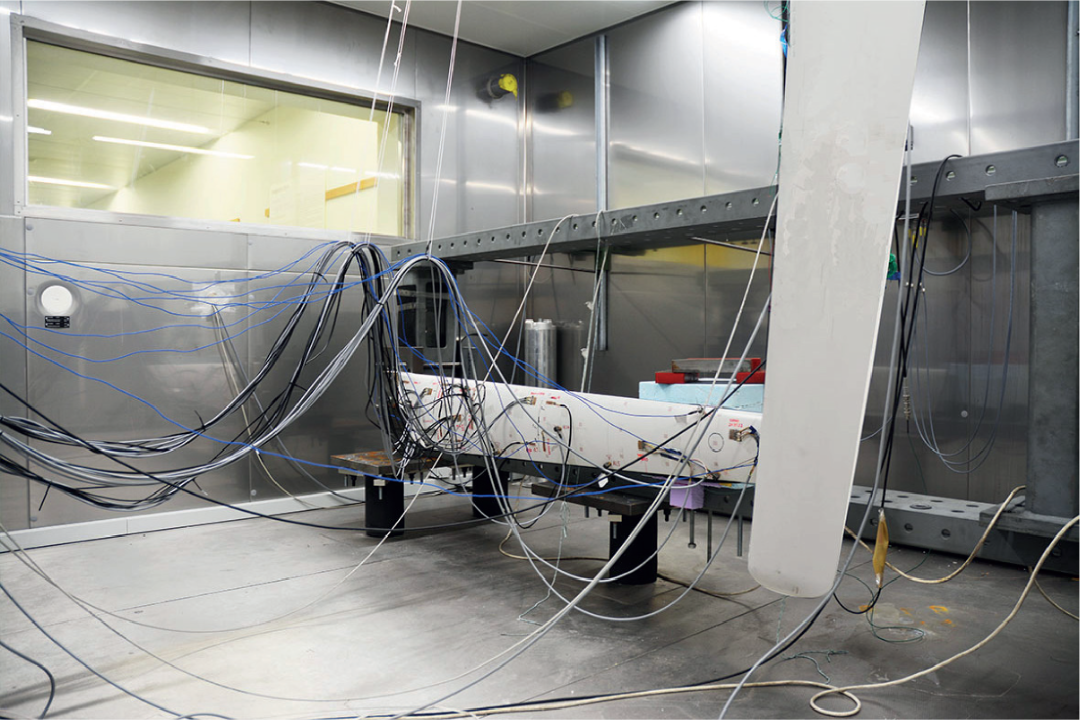}
\caption{Experiment setup of the wind turbine blade.}
\label{experiment}
\end{figure}
\begin{figure*}[h]
\centering
\includegraphics[width=0.9\linewidth]{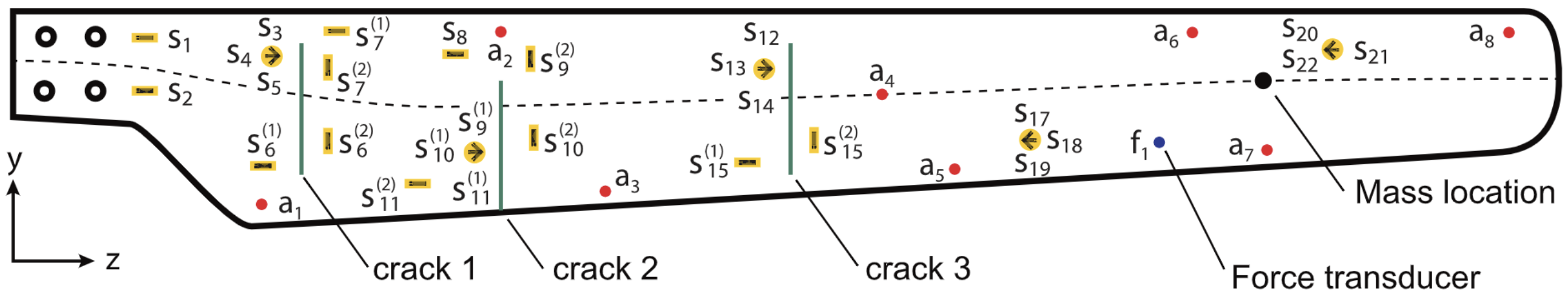}
\caption{Sensor placement of the wind turbine blade. There are totally eight accelerometers mounted on the blade, marked with red color and label $a_x$. The data from these accelerometers are the output responses, while the data from the force transducer $f_1$ are the input excitations.}
\label{WT}
\end{figure*}
\begin{figure*}[h]
\centering
\includegraphics[width=1.0\linewidth]{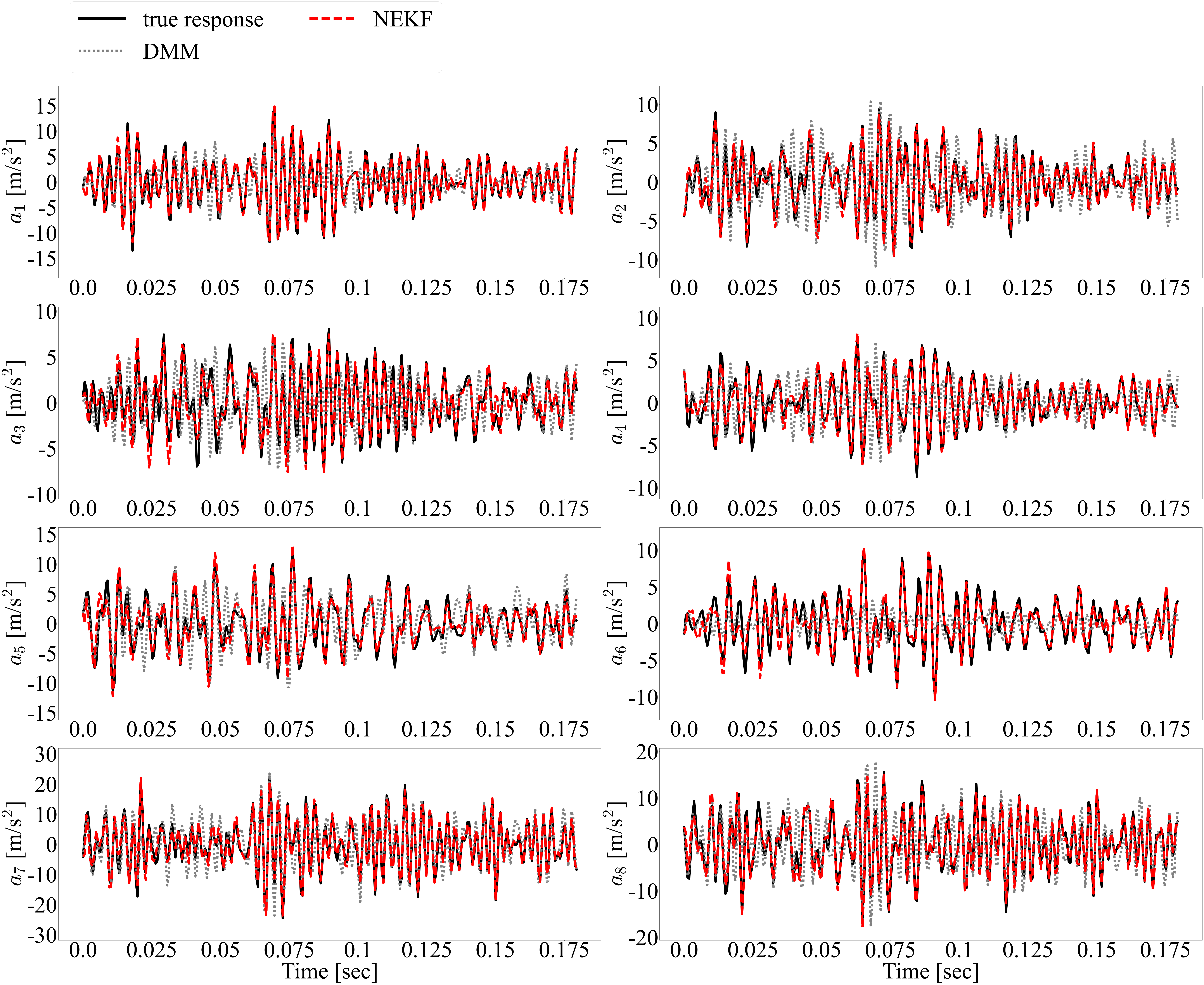}
\caption{Predicted and true responses of accelerations $a_1$ to $a_8$ for the wind turbine blade. The model is trained on a dataset of the healthy state under 25$^\circ$C and used to generate predictions on another dataset of the same state.}
\label{timeseries}
\end{figure*}
\begin{figure*}[h]
\centering
\includegraphics[width=1.0\linewidth]{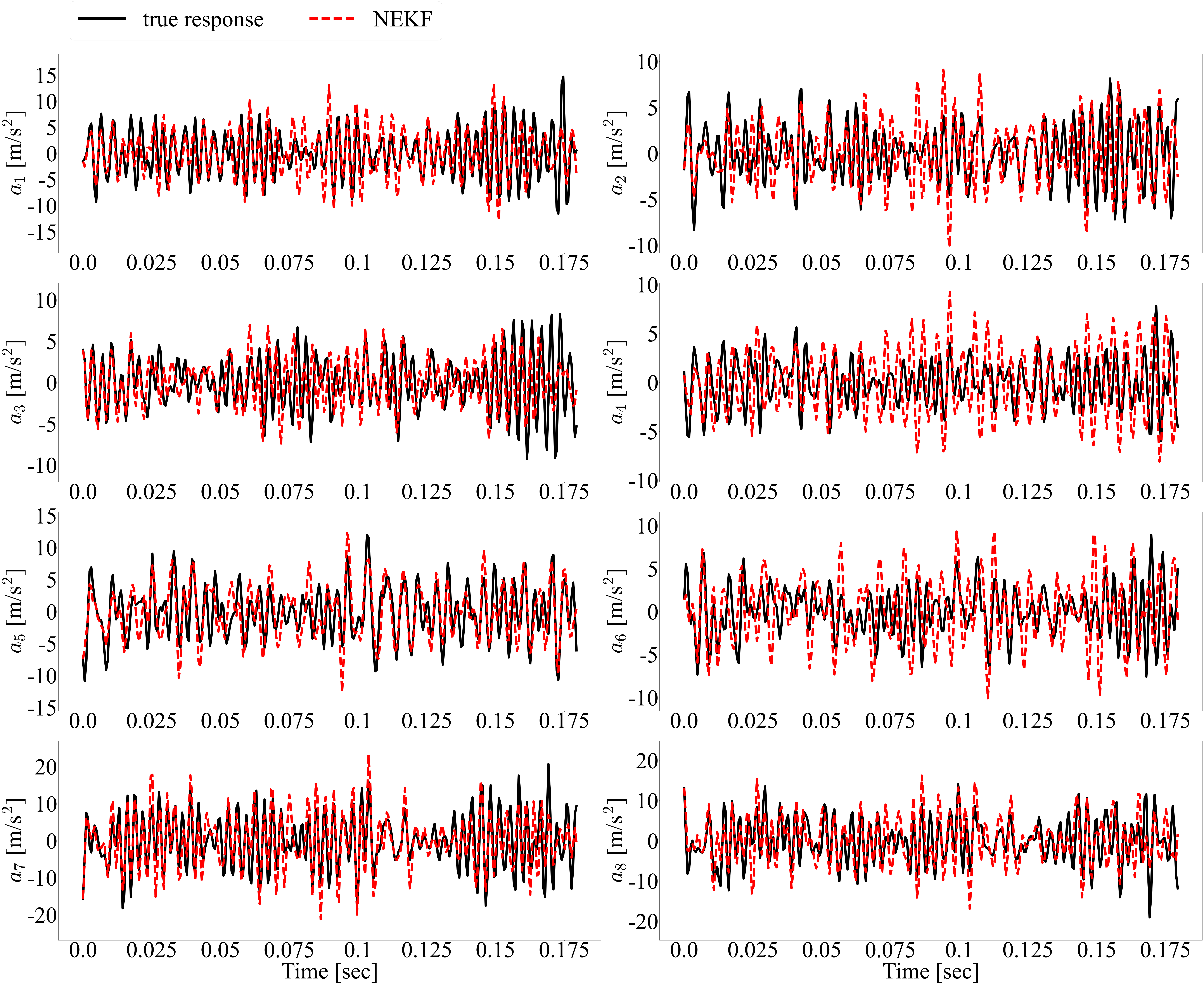}
\caption{Predicted and true responses of accelerations $a_1$ to $a_8$ for the cracked wind turbine blade. The model is trained on a dataset of the healthy state under 25$^\circ$C and used to generate predictions on a dataset of the damaged state with three 15cm cracks under 25$^\circ$C.}
\label{timeseries_L}
\end{figure*}
\subsection{Experimentally Tested Wind Turbine Blade}
Among various application fields of Structural Health Monitoring (SHM), wind turbines are gaining increased attention due to their critical significance and competitiveness as a major source of renewable energy. To further demonstrate the value of the framework in SHM for assessing performance and assisting decision making, we validate use of the Neural EKF for vibration monitoring of operational wind turbines. The data used in this paper were obtained and illustrated in \cite{ou2021vibration} by experimentally testing a small-scale wind turbine blade, as shown in Figure \ref{experiment}, in both healthy and damaged states under varying environmental temperature conditions. All the testing cases are listed in Table \ref{tab:results}. Multiple accelerometers and strain gauges are implemented on different positions of the wind turbine. We only make use of accelerometer measurements as the system outputs and the external forces measured at the force transducer $f_1$ as the system input to conduct a vibration-based assessment in this case study. The sensor configuration is shown in Figure \ref{WT}, where the red dots indicate the positions of accelerometers $a_i \quad (i= 1,2,...,8)$, with $i$ denoting the label of each sensor. The measured signals are low-pass filtered with a cut-off frequency of 380 Hz.

\subsubsection{Structural response prediction}
We train the model using the R(+25) dataset (the healthy state under 25$^\circ$C) and leverage the trained model to carry out response predictions given the corresponding forcing data for each case. Figure \ref{timeseries} shows an example of the prediction results for the case R(+25), plotted against the measured true responses. Note that the shown prediction results are based on an independent experiment different from the dataset used for training. Strong agreement between the predictions and true responses can be observed for all the eight channels, {\color{black} with an average root mean squared error (RMSE) of 1.08} . Also, the DMM is trained on the same training dataset and performs prediction on the same test dataset. The results are also plotted in gray in Figure \ref{timeseries}, {\color{black} with an average RMSE of 3.82}, and it is obvious that the conventional VAE type model fails to generate accurate predictions as Neural EKF.

\subsubsection{Anomaly detection}
\begin{figure*}
\begin{floatrow}
\capbtabbox{
\begin{tabular}{c c}
    \toprule
	Case label & Description \\
	\midrule
	R (+25) & Healthy state \\
	R (+20) & Healthy state \\
	R (-15) & Healthy state \\
	R (+40) & Healthy state \\
	A (+25) & Added mass $1\times44\text{g}$ \\
	B (+25) & Added mass $2\times44\text{g}$ \\
	C (+25) & Added mass $3\times44\text{g}$ \\
	D (+25) & $\text{Cracks}=(5\text{cm},0\text{cm},0\text{cm})$ \\
	E (+25) & $\text{Cracks}=(5\text{cm},5\text{cm},0\text{cm})$ \\
	F (+25) & $\text{Cracks}=(5\text{cm},5\text{cm},5\text{cm})$ \\
	G (+25) & $\text{Cracks}=(10\text{cm},5\text{cm},5\text{cm})$ \\
	H (+25) & $\text{Cracks}=(10\text{cm},10\text{cm},5\text{cm})$ \\
	I (+25) & $\text{Cracks}=(10\text{cm},10\text{cm},10\text{cm})$ \\
	J (+25) & $\text{Cracks}=(15\text{cm},10\text{cm},10\text{cm})$ \\
	K (+25) & $\text{Cracks}=(15\text{cm},15\text{cm},10\text{cm})$ \\
	L (+25) & $\text{Cracks}=(15\text{cm},15\text{cm},15\text{cm})$ \\
	\bottomrule
    \end{tabular}}
    {\caption{Testing Cases and prediction RMSE via Neural EKF (the model is trained using the R (+25) dataset)}\label{tab:results}}
\ffigbox{\includegraphics[width=1.1\linewidth]{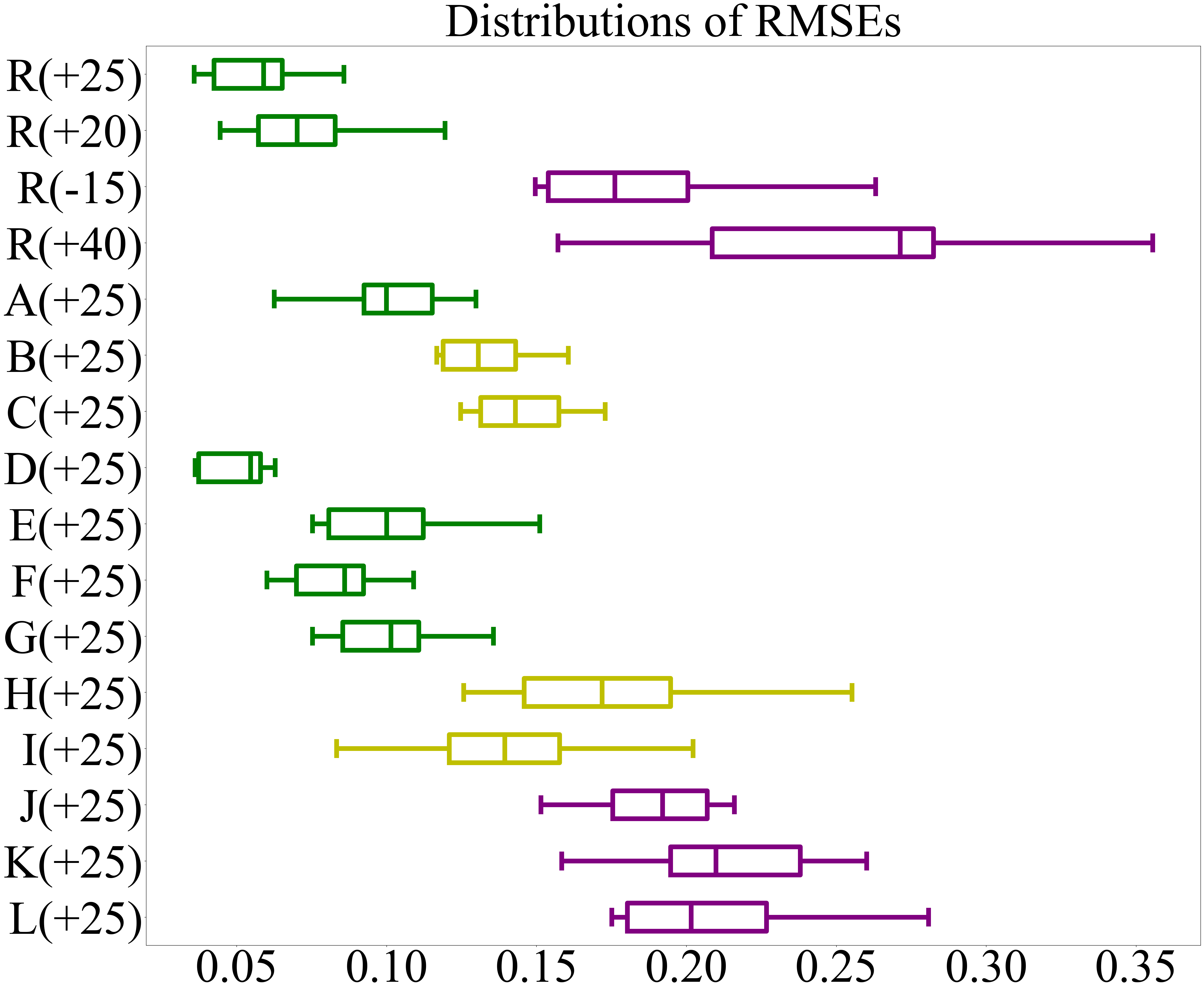}}
{\caption{Distributions of RMSEs (the color code is the same as Figure \ref{fig:PCA})}\label{fig:boxplot}}
\end{floatrow}
\end{figure*}

As stated, the model we trained uses the data from the healthy state. It is expected that for cases that deviate from the healthy baseline, such as damage, added mass, and diversified temperature conditions, the measured responses should bear discrepancies to the response predicted from the model trained on healthy data. An example of predicting the response for the case L(+25) is shown in Figure \ref{timeseries_L}, where apparent inconsistency in the response prediction is observed. We quantify the prediction errors of 8 channels for each case using RMSE, also presented in Figure \ref{fig:boxplot} using a box plot.

\begin{figure}[h]
\centering
\includegraphics[width=1.0\linewidth]{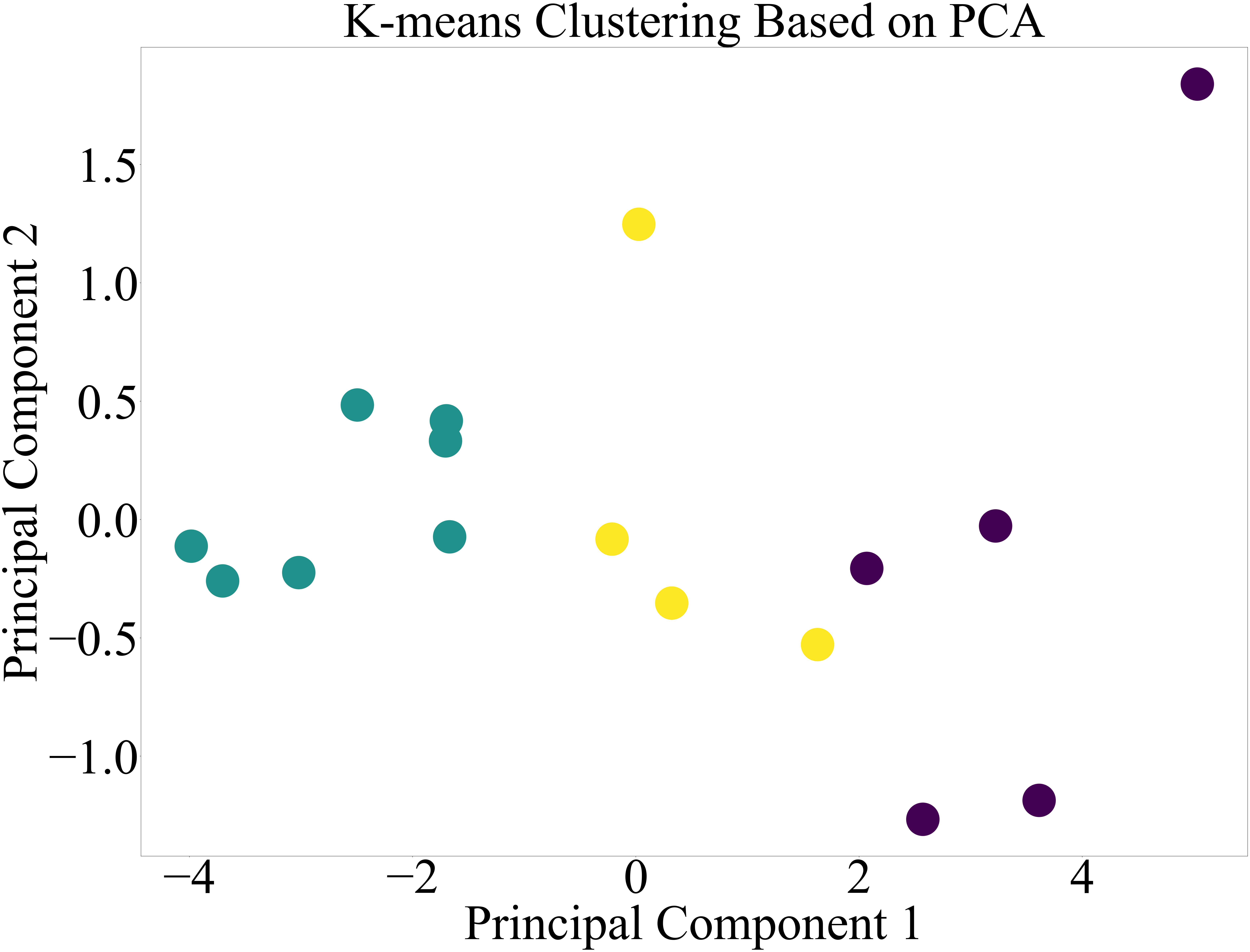}
\caption{K-means clustering results based PCA.}
\label{fig:PCA}
\end{figure}
For each case, the RMSE is 8-dimensional (as we have 8 channels of response measurement). We leverage principal component analysis (PCA) and K-means clustering, to reduce the dimension and cluster the RMSE values in terms of the first two principal components, as shown in Figure \ref{fig:PCA}. Three distinct clusters are formed, and the data points inside each cluster suggest that the RMSEs have similar features:
\begin{itemize}
    \item the first cluster in green contains Case R (+25) (baseline model), Case R (+20), Case A and Case D to Case G, which are of minor discrepancy from the baseline model\\
    \item the second cluster in yellow contains Case B, C, and Case H, I, all of which are of medium discrepancy from the baseline model;\\
    \item the third cluster in purple contains Case R (-15), Case R (+40) and Case J, K, L, all of which are of the most discrepant situations.
\end{itemize}
Notably, the clustering of RMSEs is consistent with the different degrees of anomaly. This clustering result concludes that the RMSEs are reasonable and can be used as an appropriate indicator of the anomaly degree. Here we simply use clustering to visualize the implication of RMSEs, instead of making it more quantifiable. Advanced investigation can be employed to further perform damage quantification and localization in future work.

\section{Conclusions}
In this work, we utilize a Neural Extended Kalman Filter (Neural EKF) that blends the benefits from both VAE and EKF, for data-driven modeling and response prediction of complex dynamical systems. The structure of the Extended Kalman Filters is exploited for conducting inference under the variational inference approach, which guarantees a more accurate dynamics model compared to conventional variational autoencoders. We investigate the framework on different vibration-based datasets from real-world structural/mechanical systems. The results validate the capability of the proposed Neural EKF to adequately capture the underlying dynamics of complex systems and therefore to generate accurate prediction. This is essential for downward applications, such as structural health monitoring and predictive maintenance planning. The Neural EKF-learned model offers a potent surrogate for real-world operating systems and provides the potential to establish robust structural digital twins for complex monitored systems.





\begin{dci}
The authors declared no potential conflicts of interest with respect to the research, authorship, and publication of this article.
\end{dci}

\begin{acks}
The research was conducted at the Future Resilient Systems at the Singapore-ETH Centre, which was established collaboratively between ETH Zurich and the National Research Foundation Singapore. This research is supported by the National Research Foundation Singapore (NRF) under its Campus for Research Excellence and Technological Enterprise (CREATE) programme. The project is also supported by the Stavros Niarchos Foundation through the ETH Zurich Foundation and the ETH Zurich Postdoctoral Fellowship scheme.
\end{acks}

\bibliographystyle{SageV}
\bibliography{my_references.bib}

{\color{black}
\section{Appendix}
\subsection{Latent Dynamics}
Given that both the transition and observation models are constructed using neural networks, there exist non-unique combinations of these two models that can comparably produce reasonable system responses. It is possible to achieve a good output response with incorrect dynamics and an appropriate selection of parameters $\mathbf{Q}$ and $\mathbf{R}$, which is in fact the philosophy of Kalman filters. $\mathbf{Q}$ and $\mathbf{R}$ represent the process and measurement noise covariance matrices, which are meant to account for modeling and measurement noise, but which nonetheless necessitate a fairly competent process and measurement model for successful operation. The difference in a Deep version of such a Bayesian filter (as the Neural EKF) is that the state-space equations (process and measurement equations), as well as the noise sources, are not prescribed, but learned. This is why the dynamics, as stated in the title, is actually learnable. This is a main difference with respect to standard Bayesian filters, where these parameters, including the noise covariances, need to be preset. As in conventional filtering, here as well, different combinations of transition and observation models with different $\mathbf{Q}$ and $\mathbf{R}$ matrices exist that can accurately reconstruct the response; the algorithm is searching for a local optimum that represents one of these solutions. This implies that the transition model may not be aligned with the actual dynamics of the system, but rather a nonlinear transformation of it, as the observation model is not explicitly defined.

Figure \ref{fig:phase_comparison_0.1} illustrates the rotated learned latent states by identifying an optimal linear relationship with the true system states. The rotated latent states exhibit a similar pattern to the true system states, albeit with presence of some stretching. This is because the relationship between learned and true states is not a simple linear relationship, but rather a nonlinear transformation. In contrast, the DMM-predicted latent states are plotted on the bottom and exhibit distinctly different patterns to the true system states, even after optimal rotation.

When inference of such true latent states is of interest, we suggest the use of a continuous model, such as Neural Ordinary Differential Equations (Neural ODEs$^{18}$), which allows to directly obtain the derivative of the latent state variables.
Neural ODEs parameterize the derivative of the latent state (vector field) using a neural network. The output of the network is computed using a black-box differential equation solver. Neural ODEs have proven to be particularly useful for continuous-time latent variable models and can be used to model continuous dynamics, thereby enabling the introduction of derivative relations in the latent space and structure our observation model. We do not illustrate more on this approach here, since this comprises an extension which is beyond the objective of this work, which aimed to illustrated the benefits of the EKF structure in learning dynamical systems. We note $\mathbf{z}=[z_1,z_2]$, where $z_1$ and $z_2$ represent the first and second half elements of the latent state, respectively. Similar to the formulations in Eq.(2) and Eq.(3), but now set in a continuous form with the Neural ODE modeling the transition function, the dynamics can be modeled as:
\begin{align}
\dot{z}_1 &= z_2, \\
\dot{z}_2 &= f(z_1,z_2,\mathbf{u}) + w, \\
\mathbf{x} &= f(z_1,z_2,\mathbf{u}) + v, \label{eq:acc}
\end{align}
\begin{figure}[h]
\begin{subfigure}{.45\linewidth}
\centering
\includegraphics[width=\linewidth]{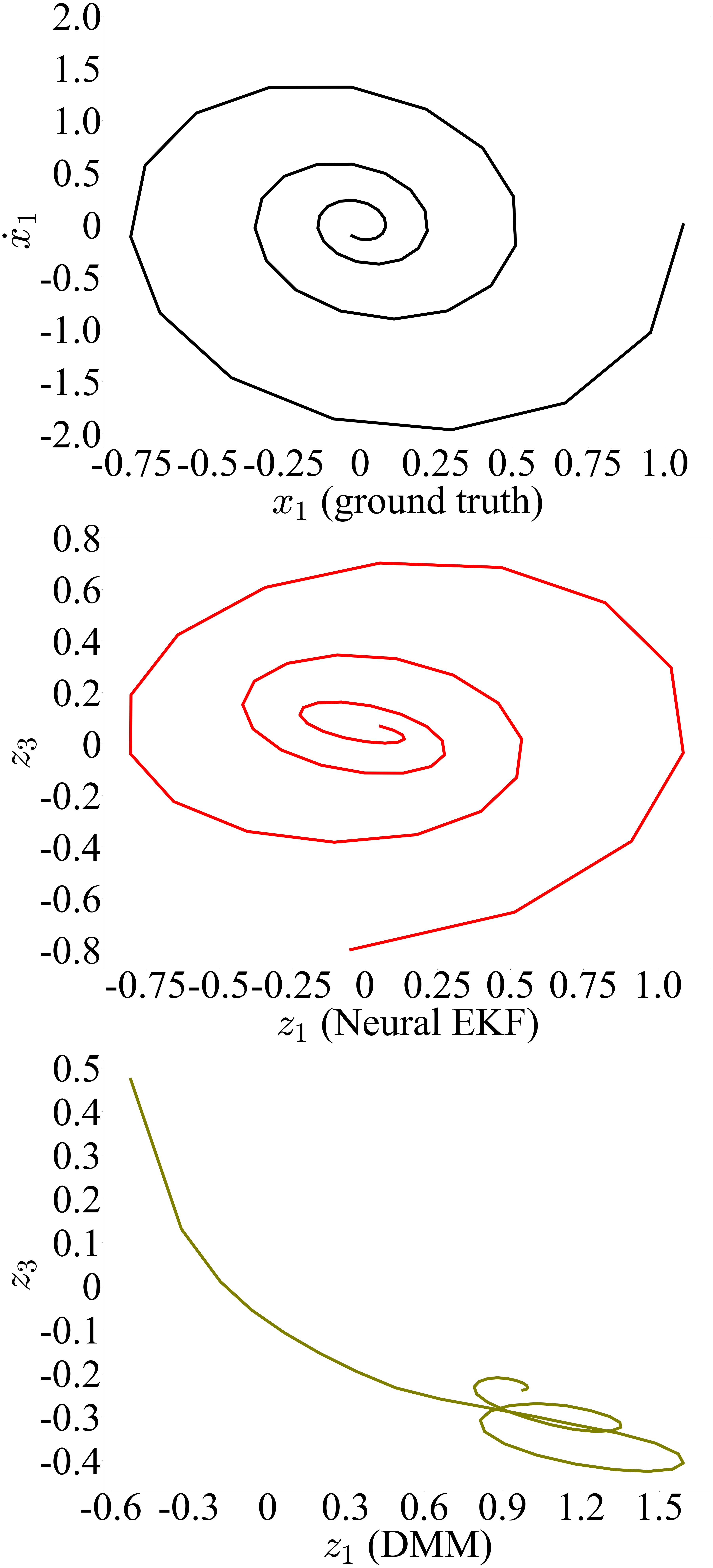}
\caption{Phase portraits of the latent states $z_1$ versus $z_3$, estimated via the Neural EKF (middle) and DMM (bottom), compared to the ground-truth displacements $x_1$ versus velocities $\dot{x}_1$ of the first DOF (top).}
\label{fig:phase_0.1}
\end{subfigure}
\begin{subfigure}{.45\linewidth}
\centering
\includegraphics[width=\linewidth]{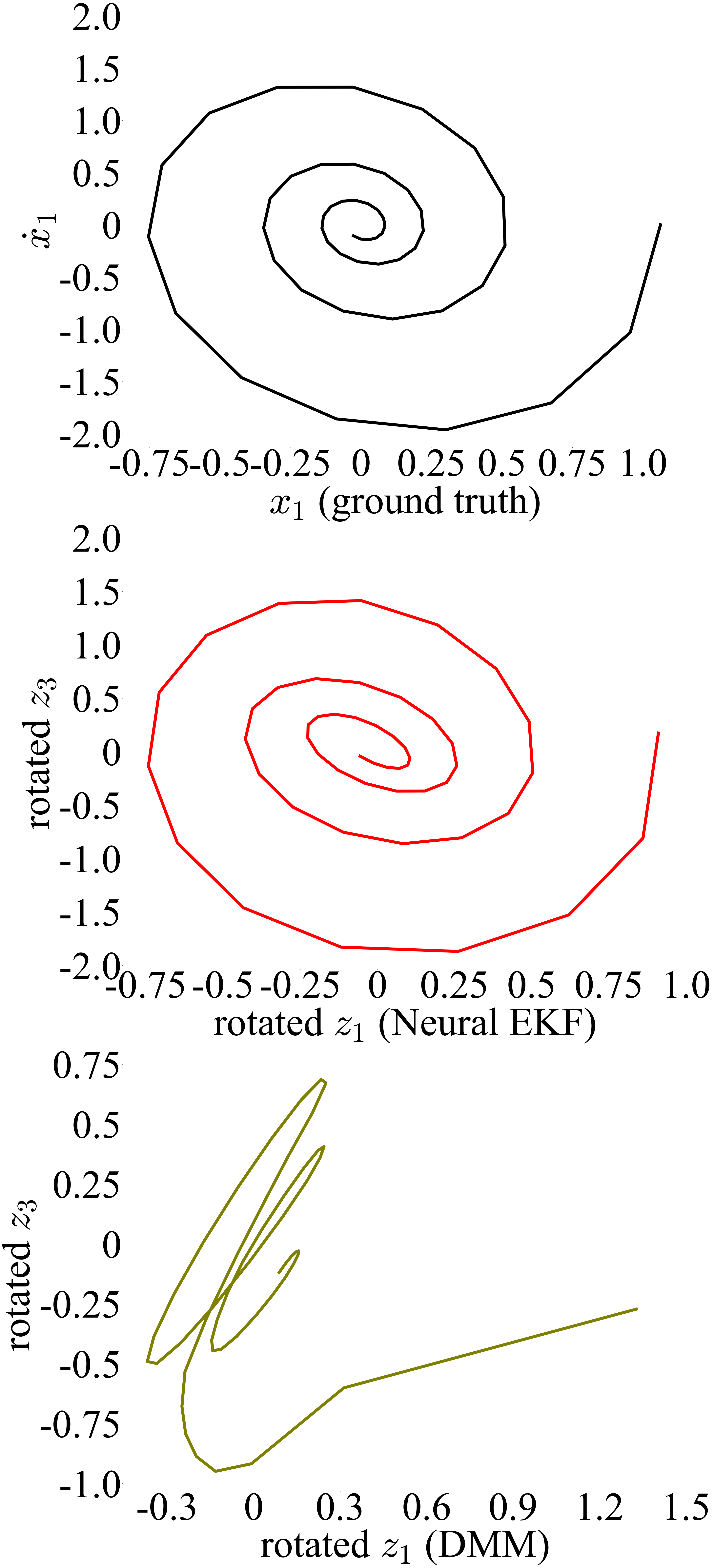}
\caption{Phase portraits of the optimally rotated latent states $z_1$ versus $z_3$, estimated via the Neural EKF (middle) and DMM (bottom), compared to the ground-truth displacements $x_1$ versus velocities $\dot{x}_1$ of the first DOF (top).}
\label{fig:phase_0.1_rot}
\end{subfigure}
\caption{Comparison of the latent states estimated via the Neural EKF and DMM against the ground-truth phase portraits. The results are obtained on the Duffing system, which has been simulated in Section \textit{Duffing Oscillator System} with observation noise variance of 0.1.}
\label{fig:phase_comparison_0.1}
\end{figure}

\begin{table*}[h]
\caption{Estimated covariance matrices $\mathbf{Q}$ and $\mathbf{R}$ under different noise levels} 
\centering 
\begin{tabular}{c cccccc} 
\toprule
& \multicolumn{4}{c}{Transition process} &\multicolumn{2}{c}{Observation process} \\
\cmidrule(lr){2-5} \cmidrule(lr){6-7}
Noise variance &$q_1$ &$q_2$ &$q_3$ &$q_4$ &$r_1$ &$r_2$\\
\midrule
0.001 & 0.000 & 0.7886e-26 & 4.9305e-34 & 6.7735e-25 & 3.3331e-9 & 6.8190e-8 \\  
0.01 & 9.4104e-35 & 2.8987e-14 & 5.4440e-15 & 1.1494e-27 & 2.4185e-5 & 2.7647e-5 \\  
0.1 & 6.0510e-31 & 6.3190e-09 & 2.6425e-11 & 1.6114e-19 & 0.0030 & 0.0029 \\  
\bottomrule
\end{tabular}
\label{tab:covariance}
\end{table*}
}

The first set of equations serves for enforcing the second half of the latent state to assume the form of the derivatives of the first half, while the second equation models the dynamics. Again, this is only when it is of interest to specifically infer a physical latent space. Since the second set of equations is usually chosen to generate a system's accelerations (as derived from a physics-based equation of motion), it is possible in this case to eliminate the requirement for an additional neural network, for example $g$ in Eq.(3), for modeling the observation process, and the observation function is merely$f$ in this case, as indicated in Eq.\eqref{eq:acc}. However, in this case the purpose of the network is modified, since the idea here is to impose some sort of physics-based structure, as opposed to more freely learn the unknown underlying dynamics.
\subsection{Discussion on Noises}
As described in the theoretical section, the covariance matrices $\mathbf{Q}$ and $\mathbf{R}$ are assumed diagonal, thus for the duffing oscillator example, these are represented by $\mathbf{Q}=\text{diag}(q_1,q_2,q_3,q_4)$ and $\mathbf{R}=\text{diag}(r_1,r_2)$, respectively. The estimated diagonal elements are listed in Table \ref{tab:covariance}. 

It is observed that 1) the estimated values of $\mathbf{Q}$ are close to zero and much smaller than the defined values of $\mathbf{R}$, as no process noise was in this example added to the simulated data. 2) Without structuring the observation model, both transition and observation models are parameterized by neural networks, jointly optimized to reconstruct system response. Thus, it is not surprising  that the estimated $\mathbf{Q}$ and $\mathbf{R}$ are not exactly matching the true value, but one can still observe that the estimated values of $\mathbf{R}$ increase when larger observation variances are added to the simulated data, thereby accounting for increasing uncertainty in the measured data. It is important to underline that, unlike the conventional EKF, the proposed Neural EKF is a deep learning-based method, meant to learn (infer) the transition and emission models from data. The results in Figure \ref{fig:noisy_comparison} show that even with highly noisy data, Neural EKF is able to recover the true response with high confidence, thus the estimated $\mathbf{R}$ is observed to be smaller. Using the estimated standard deviations, which are the square roots of the estimated variances, 99.7\% confidence intervals for predicted responses can be calculated. These intervals should be highly precise around mean values, considering the extremely low values of estimated variances.



\end{document}